# Visual-Inertial SLAM for Unstructured Outdoor Environments: Benchmarking the Benefits and Computational Costs of Loop Closing

Fabian Schmidt[1,2] | Constantin Blessing[1] | Markus Enzweiler[1] | Abhinav Valada[2]

[1]Institute for Intelligent Systems, Esslingen University of Applied Sciences, Esslingen, Germany

[2]Department of Computer Science, University of Freiburg, Freiburg, Germany

**Correspondence**
Corresponding author Fabian Schmidt.
Email: fabian.schmidt@hs-esslingen.de

## Abstract

Simultaneous Localization and Mapping (SLAM) is essential for mobile robotics, enabling autonomous navigation in dynamic, unstructured outdoor environments without relying on external positioning systems. These environments pose significant challenges due to variable lighting, weather conditions, and complex terrain. Visual-Inertial SLAM has emerged as a promising solution for robust localization under such conditions. This paper benchmarks several open-source Visual-Inertial SLAM systems, including traditional methods (ORB-SLAM3, VINS-Fusion, OpenVINS, Kimera, and SVO Pro) and learning-based approaches (HFNet-SLAM, AirSLAM), to evaluate their performance in unstructured natural outdoor settings. We focus on the impact of loop closing on localization accuracy and computational demands, providing a comprehensive analysis of these systems' effectiveness in real-world environments and especially their application to embedded systems in outdoor robotics. Our contributions further include an assessment of varying frame rates on localization accuracy and computational load. The findings highlight the importance of loop closing in improving localization accuracy while managing computational resources efficiently, offering valuable insights for optimizing Visual-Inertial SLAM systems for practical outdoor applications in mobile robotics. The dataset and the benchmark code are available under `https://github.com/iis-esslingen/vi-slam_lc_benchmark`.

## 1 Introduction

Simultaneous Localization and Mapping (SLAM) is a critical technology in the field of mobile robotics. SLAM enables robots to build a map of an unknown environment while simultaneously keeping track of their current location within that map. The primary purpose of SLAM is to provide autonomous robots with the ability to navigate and operate effectively in dynamic (Bešić & Valada, 2022) and unstructured environments (Capua et al., 2018; Cremona et al., 2022; Pire et al., 2019) without relying on external positioning systems. Many SLAM systems employed in outdoor applications depend on Global Navigation Satellite System (GNSS), with RTK-GNSS (Real-Time Kinematic Global Navigation Satellite System) being particularly costly and requiring extensive base station coverage to ensure accuracy (Aguiar et al., 2020). Alternative solutions leverage consumer-grade GNSS, integrating its output with various onboard sensors to improve localization accuracy (Cao et al., 2022; Cattaneo & Valada, 2024; Nayak et al., 2023; Arce et al., 2023). However, GNSS signals are not consistently available and may be unreliable due to environmental conditions, such as dense tree canopies in orchards, which can hinder signal reception and lead to catastrophic failures. Consequently, Visual-Inertial SLAM systems have facilitated numerous applications in robotics, particularly in outdoor scenarios where GNSS reliability is compromised (Cadena et al., 2016; Gosala et al., 2023).

Outdoor scenarios, unlike urban environments, face harsh environmental conditions, e.g., seasonal changes, drastic illumination changes, large open fields, and irregular terrain, making navigation more difficult. The ability of Visual-Inertial SLAM systems to autonomously construct and update maps of their surroundings while simultaneously determining their own position has profoundly influenced the field of autonomous navigation in outdoor scenarios. However, when estimating the robot's motion over time, Visual-Inertial SLAM leads to drift, challenging practical applications. For long-duration operations, a consistent and robust system is required that demonstrates adequate resistance to drift (Angeli et al., 2008; Vödisch et al., 2022, 2023).

In Visual-Inertial SLAM systems, loop closing (LC) is a key technique to minimize drift by identifying previously visited locations. This technique corrects the robot's historical trajectory and optimizes the map estimation, enhancing both localization accuracy and map integrity over time. LC helps maintain precise navigation and mapping, which is essential for autonomous ground vehicles performing systematic coverage tasks, such as terrain monitoring, maintenance, and vegetation management (Islam et al., 2023). By identifying loops, the system can adjust the map and the robot's position to mitigate cumulative errors. While theoretically promising, LC introduces practical challenges. It requires the system to continuously compare current observations with past data to detect overlaps, demanding substantial computational power. This demand is particularly challenging for mobile robots, where computational resources are limited, and real-time processing capability is crucial. Efficiently managing these demands is essential for maintaining accurate localization without compromising performance or battery life (Zhao et al., 2020). Robust algorithms that minimize computational overhead are critical for practical applications of Visual-Inertial SLAM in mobile robotics (Tsintotas et al., 2022).

In previous research (Schmidt, Holzmüller, et al., 2024), we benchmarked several open-source Visual-Inertial SLAM systems in unstructured outdoor settings, evaluated the impact of LC on localization accuracy, and analyzed the computational demands. In this study, we expand on our results, conducting experiments to assess how varying frame rates influence localization accuracy and computational load. In addition, we assess several newly integrated methods, including Kimera (Rosinol et al., 2021), SVO Pro (Forster et al., 2017), HFNet-SLAM (Liu & Aitken, 2023), AirSLAM (Xu et al., 2025), and HLoc (Sarlin et al., 2019) as an LC extension for OpenVINS (Geneva et al., 2020), providing insights into optimizing Visual-Inertial SLAM for practical use. Our contributions include:

1. **Benchmarking Open-Source Visual-Inertial SLAM Methods**: We conduct a comprehensive evaluation of the application of various open-source Visual-Inertial SLAM methods in unstructured outdoor environments, focusing on their performance and applicability in natural settings.
2. **Quantitative and Qualitative Analysis of LC Effects**: We evaluate the impact of LC on localization accuracy across diverse driving scenarios and environmental conditions, providing a detailed analysis of its benefits and limitations.
3. **Resource Analysis**: We investigate the additional computational overhead imposed by LC, highlighting the trade-offs in computational resources.
4. **Frame Rate Impact Study**: We research the impact of different frame rates on localization accuracy and computational effort, offering insights into optimizing Visual-Inertial SLAM systems for real-time applications.

The remainder of this paper is structured as follows: Section 2 reviews existing Visual and Visual-Inertial SLAM benchmarks across various environments and settings. Section 3 describes the experimental setup, detailing the specific Visual-Inertial SLAM methods used, the dataset, and the evaluation criteria. Section 4 presents the results and discusses their implications and relevance for autonomous navigation. Finally, Section 5 concludes the paper by summarizing key findings and proposing future research directions.

## 2 Related Work

The extensive literature on SLAM systems provides insights into applications, difficulties, and performance in a variety of scenarios, highlighting their significance in the development of robotic navigation. This chapter focuses on benchmarks for Visual and Visual-Inertial SLAM, essential for the evaluation and comparison of individual SLAM performances.

Numerous benchmarks (Buyval et al., 2017; Ibragimov & Afanasyev, 2017; Filipenko & Afanasyev, 2018; Delmerico & Scaramuzza, 2018; Giubilato et al., 2018, 2019; Mingachev et al., 2020; Gao et al., 2020; Servières et al., 2021; Merzlyakov & Macenski, 2021; Bahnam et al., 2021; Bujanca et al., 2021; Herrera-Granda et al., 2023; Sharafutdinov et al., 2023; Passalis et al., 2022) systematically evaluate the localization accuracy of contemporary Visual and Visual-Inertial SLAM algorithms through the use of renowned datasets such as TUM RGB-D (Sturm et al., 2012), EuRoC (Burri et al., 2016), and KITTI (Geiger et al., 2012). These benchmarks offer insights into the efficacy of SLAM algorithms across diverse environments, from indoor spaces to unpredictable outdoor settings, incorporating various robotic platforms, such as drones, mobile robots, and cars. Furthermore, several benchmarks, such as (Delmerico & Scaramuzza, 2018), (Bahnam et al., 2021), (Giubilato et al., 2018), (Giubilato et al., 2019), (Bujanca et al., 2021), and (Sharafutdinov et al., 2023), evaluate the computational performance of SLAM algorithms in addition to localization accuracy. (Delmerico & Scaramuzza, 2018) specifically monitor CPU and RAM usage across different embedded computing platforms. (Giubilato et al., 2018, 2019) evaluated stereo Visual SLAM systems on an embedded platform, both with and without GPU support, whereas (Bahnam et al., 2021) provide detailed statistics on computational timings. In contrast, (Bujanca et al., 2021) focus on ensuring consistent localization accuracy results across various platforms, including an embedded platform, a consumer-grade laptop, and a high-end workstation. (Sharafutdinov et al., 2023) conducted an extensive evaluation of 13 Visual and Visual-Inertial SLAM algorithms, focusing on their performance and reliability in autonomous navigation for both indoor and urban outdoor scenarios, while also monitoring memory usage and CPU utilization.

In addition to common benchmarks, studies such as (Chahine & Pradalier, 2018; Joshi et al., 2019; Tschopp et al., 2019; R. Li et al., 2023) specifically explore Visual and Visual-Inertial SLAM applications in particular contexts, demonstrating their adaptability and potential constraints in challenging environments. (Chahine & Pradalier, 2018) utilize the Symphony Lake dataset (Griffith et al., 2017), which consists of images of a lake shore captured over three years across different seasons. They evaluated three different SLAM methods to assess the seasonal influence on localization accuracy. (Joshi et al., 2019) evaluate ten open-source Visual-Inertial SLAM algorithms in marine settings, addressing challenges such as low visibility and dynamic lighting conditions. This comprehensive analysis, using datasets from underwater robots, provides insights into the performance of direct and feature-based SLAM methods. (Tschopp et al., 2019) explore Visual Odometry methods for rail vehicles, emphasizing the combination of stereo vision with inertial measurement units (IMU) to improve the precision of motion estimation. They highlight the importance of robust algorithms that address challenges such as high velocities and constrained motion and compare various Visual-Inertial Odometry frameworks to showcase their strengths and limitations. (R. Li et al., 2023) explore the application feasibility of monocular Visual-Inertial SLAM methods in freight railways. They emphasize challenges including scale estimation errors and frequent failures attributed to the constrained motion patterns.

Benchmarks in agricultural environments highlight the specific challenges of implementing SLAM technologies, which must adapt to dynamic conditions and complex landscapes. (Capua et al., 2018) study the application of Visual SLAM in orange orchards, highlighting the need for systems capable of adapting to the diverse and cluttered environments typical for agriculture. (Hroob et al., 2021) evaluate Visual SLAM systems in a simulated vineyard, demonstrating SLAM's potential in precision agriculture. Their study focuses on the challenges posed by the vineyard's structured yet dynamic environment, including varying vine heights and the influence of changing sunlight throughout the day. (Comelli et al., 2019) and the follow-up (Cremona et al., 2022) cover state-of-the-art stereo Visual-Inertial SLAM systems on an arable farming dataset (Pire et al., 2019), specifically in soybean fields. These studies highlight the operational challenges faced by using SLAM in agricultural contexts, evaluating their accuracy and robustness. They also address the adaptability of SLAM systems to the distinct characteristics of arable farming, such as crop height variability and environmental factors such as wind and lighting conditions.

In this paper, we aim to address key gaps in the existing literature by focusing on unstructured outdoor environments such as gardens or parks, which have been underrepresented in current benchmarks. These environments present distinct challenges for SLAM technologies, including highly variable vegetation density, uneven terrain, and fluctuating lighting conditions due to canopy coverage. Unlike traditional natural settings that have been explored in existing benchmarks (Capua et al., 2018; Hroob et al., 2021; Comelli et al., 2019; Cremona et al., 2022), which often deal with structured row crops and controlled variables, gardens and parks offer a more dynamic and unpredictable landscape. This necessitates the development of more adaptable and robust SLAM algorithms capable of handling sudden environmental changes and less predictable pathways. Furthermore, our approach includes a detailed study of the interplay between environmental complexity and SLAM system performance, providing insights into how these factors influence the operational efficacy of robotic navigation. By exploring these unique settings, our research contributes to a broader understanding of SLAM applicability and enhances the technology's adaptability to diverse natural environments.

# 3 Experimental Setup

In this section, we first describe the dataset that we use for benchmarking. Second, we introduce the selected Visual-Inertial SLAM algorithms chosen for evaluation on the aforementioned dataset. Finally, we describe the evaluation methodology in more detail.

## 3.1 Dataset

For data recording, we utilized the unmanned ground vehicle (UGV) described in (Schmidt, Blessing, et al., 2024), which provided the necessary mobility and versatility to navigate the environments effectively. The UGV was equipped with an Intel RealSense T265 camera, augmented with a six degrees of freedom (DOF) IMU. The camera captures imagery at a frequency of 30 Hz with a resolution of 848x800 pixels, employing a monochrome global shutter and a fisheye lens with a 160° diagonal field of view. The IMU records data with the accelerometer at 62.5 Hz and the gyroscope at 200 Hz. The accuracy and reliability of data in visual localization studies are critically dependent on the quality of the ground truth data. To ensure high-precision ground truth for the robot's positioning, a Leica TS-16 Total Station was employed. It captures the robot's position using a prism mounted firmly on the robot's sensor board. The total station records three DOF positional ground truth data at a frequency of 2-5 Hz, which is sufficient given the UGV's speed of approximately 1 km/h. This setup achieves millimeter-level positional accuracy, recording a trajectory as timestamped 3D position data points.

We recorded 8 sequences, focusing on four unstructured outdoor environments that encompass diverse garden sizes and a park-like expanse, as depicted in Figure 1. The locations vary in size, ranging from 75 sqm to 500 sqm, providing a mix of small, medium, and large-scale environments. As shown in Table 1, the dataset captures different seasonal conditions, including summer, autumn, and winter, along with a range of weather scenarios such as sunny, cloudy, and windy conditions, ensuring diverse environmental challenges for evaluation. Additionally, we considered different driving scenarios that can be divided into two categories: *Perimeter* and *Lane*. In the *Perimeter* scenario, the robot completes three full rounds starting from the edge of the lawn area, ensuring that the boundaries of the area are clearly defined and consistently maintained throughout the operation. Conversely, in the *Lane* scenario, the robot moves in parallel lanes across the lawn, incorporating multiple 180° turns, resembling a methodical, straight-line path. This approach is efficient for large, regular-shaped lawns, providing uniform coverage and efficient use of time and resources.

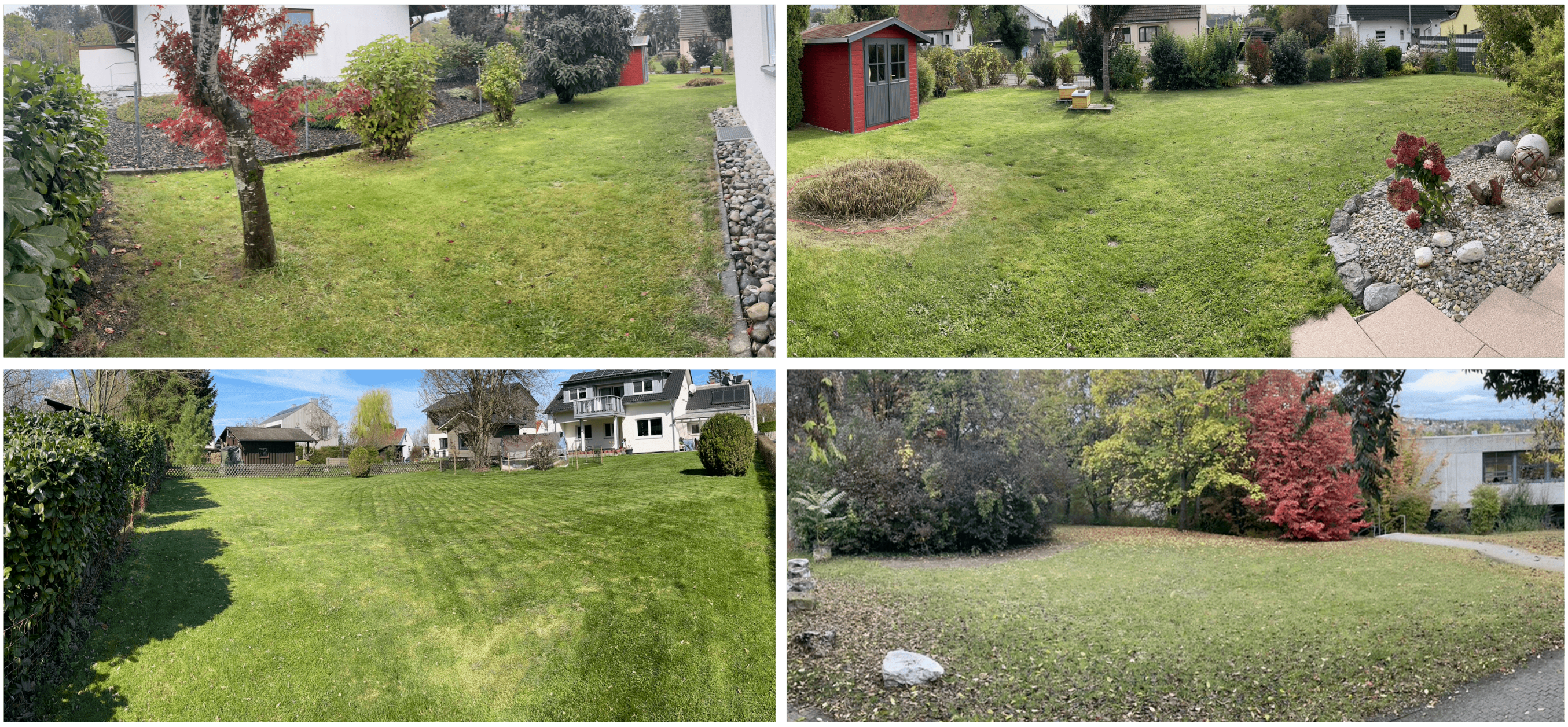

**FIGURE 1** Different dataset recording environments. Top left shows the *Garden Small*, top right *Garden Medium*, bottom left *Garden Large* and bottom right *Park* location.

**TABLE 1** Dataset characteristics.

| Location | Size [sqm] | Recording Date | Season | Weather | Scenario | Duration [s] | Distance [m] |
|---|---|---|---|---|---|---|---|
| Garden Small | 75 | 2023-12-21 - 10:04:19 | Winter | Cloudy & Windy | Lane<br>Perimeter | 189.0<br>361.1 | 45.2<br>125.6 |
| Garden Medium | 250 | 2023-09-15 - 14:55:44 | Autumn | Cloudy | Lane<br>Perimeter | 264.3<br>466.6 | 89.0<br>167.4 |
| Garden Large | 500 | 2023-08-22 - 07:29:53 | Summer | Sunny | Lane<br>Perimeter | 352.8<br>788.1 | 123.4<br>299.2 |
| Park | 250 | 2023-07-31 - 09:28:49 | Summer | Cloudy | Lane<br>Perimeter | 210.2<br>437.9 | 43.4<br>164.2 |

By covering these scenarios, we aim to provide the data to identify the effect of driving strategy on localization accuracy. Importantly, each scenario introduces a different number of rotations and movements that can potentially affect the precision of the robot's localization system. Understanding the impact of these variables is crucial for optimizing the robot's navigation and operational efficiency.

## 3.2 Algorithms

For this benchmark, we exclusively focus on open-source Visual-Inertial SLAM algorithms that utilize monocular or stereo cameras and the ROS (Robot Operating System) framework, containing an LC module. As shown in Table 2, we selected methods that are most diverse regarding their techniques in the frontend, backend, and LC to gain a comprehensive understanding of their influence on performance. A key component of SLAM is LC, which ensures global consistency by detecting previously visited locations and correcting the trajectory accordingly. The effectiveness of LC depends on the approach used for loop detection and loop correction, both of which vary across different SLAM methods. Loop detection methods rely either on bag-of-words (DBoW2) (Gálvez-López & Tardós, 2012) or global descriptor matching. DBoW2 represents images as a set of visual words based on handcrafted or learned features and organizes them into a hierarchical vocabulary, making it efficient but sensitive to viewpoint and illumination changes. Instead of relying on distinct features, learning-based approaches, i.e., NetVLAD (Arandjelovic et al., 2016), extract global descriptors from images to improve robustness to environmental variations at the cost of higher computational requirements. Both approaches incorporate a geometric verification (GV) step to refine LC candidates to reduce false positives (Tsintotas et al., 2022). For loop correction, pose graph optimization (PGO) efficiently adjusts keyframe poses within a graph-based trajectory representation, minimizing accumulated drift. However, unlike bundle adjustment (BA), PGO primarily optimizes the relative transformations between keyframes and does not directly refine the 3D structure of past keyframes and detected landmarks. Global BA, on the other hand, jointly optimizes all keyframes and landmarks, resulting in a globally consistent trajectory and map but at a significantly higher computational cost. These distinctions in loop detection and correction impact the robustness of SLAM methods, particularly in environments with seasonal changes, dynamic elements, or varying lighting conditions.

**TABLE 2** Overview of the Visual-Inertial SLAM methods and their algorithmic characteristics.

| Method | Frontend | Backend | Loop Detection | Loop Correction |
|---|---|---|---|---|
| ORB-SLAM3 | Feature (ORB) | Optimization (Local BA) | DBoW2 (ORB) + GV | PGO + Global BA |
| HFNet-SLAM | Feature (SuperPoint) | Optimization (Local BA) | Global Descriptor (NetVLAD) + GV | PGO + Global BA |
| VINS-Fusion | Feature (Shi Tomasi + KLT) | Optimization (Local BA) | DBoW2 (BRIEF) + GV | PGO |
| OpenVINS<br>+ VINS-Fusion<br>+ Maplab<br>+ HLoc | Feature (Shi Tomasi + KLT) | Filter (MSCKF) | None<br>DBoW2 (BRIEF) + GV<br>DBoW2 (BRISK) + GV<br>Global Descriptor (NetVLAD) + GV | None<br>PGO<br>PGO + Global BA<br>PGO |
| Kimera | Feature (Shi Tomasi + KLT) | Optimization (Local BA) | DBoW2 (ORB) + GV | PGO |
| SVO Pro | Semi-Direct | Optimization (Local BA) | DBoW2 (ORB) + GV | PGO / Global BA |
| AirSLAM | Feature (PLNet) | Optimization (Local BA) | DBoW2 (PLNet) + GV | PGO + Global BA |

**ORB-SLAM3** (Campos et al., 2021) stands out as a multimodal feature- and optimization-based approach with an integrated LC mechanism. The system operates through three threads: tracking, local mapping, and loop and map merging. In the frontend, ORB-SLAM3 uses a feature-based approach by extracting ORB features (Rublee et al., 2011). The tracking thread determines the current frame's pose by minimizing reprojection errors with ORB features and selects keyframes. Optimization-based methods are employed in the backend. The local mapping thread enhances the map by adjusting keyframes locally using local BA. Finally, the loop and map merging thread identifies revisited areas using a DBoW2-based keyframe database and performs LC to ensure trajectory and map accuracy by applying PGO followed by global BA.

**HFNet-SLAM** (Liu & Aitken, 2023) builds upon the ORB-SLAM3 framework but replaces traditional hand-crafted features and DBoW2-based loop detection with deep learning-based local and global feature extraction. In the tracking thread, HFNet-SLAM employs MobileNetV2 (Sandler et al., 2018) combined with SuperPoint (DeTone et al., 2018) to extract local features, improving robustness in challenging environments. This approach addresses the limitations of traditional features in challenging conditions, such as sparse textures or dynamic changes. In the mapping thread, HFNet-SLAM employs MobileNetV2 with NetVLAD to extract global features, using similarity scores between global descriptors to identify previously visited locations. As in ORB-SLAM3, loop correction is performed using PGO followed by global BA to ensure trajectory consistency and map accuracy.

**VINS-Fusion** (Qin et al., 2019) offers a versatile sensor fusion framework that leverages both visual and inertial cues for state estimation. Employing a feature-based approach in the frontend, VINS-Fusion detects visual features using the Shi-Tomasi detector (Shi & Tomasi, 1994) and tracks them with the Kanade-Lucas-Tomasi (KLT) (Lucas & Kanade, 1981) algorithm. The backend integrates an optimization-based state estimation module, which calculates the device's pose by fusing data from cameras and IMUs, ensuring accurate trajectory and orientation determination. The mapping module further refines this information by integrating environmental features and enhancing spatial awareness. VINS-Fusion includes an LC module that identifies previously visited locations using DBoW2 based on BRIEF (Calonder et al., 2010) feature descriptors and applies PGO to minimize drift over time.

**OpenVINS** (Geneva et al., 2020) is a filter-based method for visual-inertial navigation, utilizing tightly-coupled integration of camera and IMU data for precise state estimation. In the front end, it tracks visual features from camera images using a feature-based approach. The backend incorporates an Multi-State Constrained Kalman Filter (MSCKF) (Mourikis & Roumeliotis, 2007) and sliding window optimization to efficiently handle high-frequency IMU data and visual inputs for accurate updates to poses and velocities. While OpenVINS does not include LC by default, its modular design allows for the integration of the LC module from VINS-Fusion. Maplab (Cramariuc et al., 2022) can enhance OpenVINS by adding offline LC capabilities. To detect LC candidates, Maplab requires the extraction of BRISK (Leutenegger et al., 2011) features, which are then used within DBoW2 for loop detection. The maps and states estimated by OpenVINS are post-processed and further refined using PGO and global BA. Additionally, OpenVINS can be extended with HLoc (Sarlin et al., 2019), presented by (J. Li, 2023). It first performs global place recognition using NetVLAD, retrieving candidate locations, and then refines by extracting and matching local features with SuperPoint and SuperGlue (Sarlin et al., 2020) for geometric verification. This hierarchical approach significantly reduces computational overhead, making the system suitable for real-time loop closure detection and therefore is able to run online. For loop correction, PGO is applied to refine the trajectory and improve global consistency.

**Kimera** (Rosinol et al., 2021) provides a comprehensive framework that combines Visual-Inertial Odometry with metric-semantic mapping for enhanced scene understanding. Kimera uses a feature-based approach in the front end, extracting and matching visual features from camera images. The backend relies on optimization-based methods to refine pose estimates and construct a semantically rich map of the environment. The Visual-Inertial Odometry module estimates the device's trajectory by tightly integrating visual and inertial data, while the metric-semantic mapping module improves spatial awareness by incorporating geometric and semantic information. Kimera also features an LC module that identifies previously visited locations using DBoW2 based on ORB feature descriptors for loop detection and applies PGO for loop correction.

**SVO Pro** (Forster et al., 2017) is a flexible framework for visual-inertial navigation, capable of operating in both monocular and stereo-inertial configurations. The system uses a semi-direct approach in the frontend, minimizing photometric error from image intensities and incorporating feature extraction for robust pose estimation. A probabilistic depth estimation algorithm ensures efficient tracking of pixels on weak corners and edges. In the backend, the estimated poses are optimized for accuracy and consistency using a sliding-window-based local BA. SVO Pro utilizes DBoW2 with additionally extracted ORB features for loop detection and supports both PGO and BA. However, since BA is not robustly implemented for stereo-inertial configurations, only PGO is used in this work.

**AirSLAM** (Xu et al., 2025) is a Visual-Inertial SLAM system designed to enhance robustness in low-texture and repetitive environments. In the frontend, it introduces PLNet, a learning-based feature extractor that simultaneously detects keypoints and structural lines. LightGlue (Lindenberger et al., 2023) is used for feature matching, establishing correspondences between frames for consecutive pose estimation. The backend employs an optimization-based approach, utilizing local BA to refine the trajectory within a sliding window. For loop detection, AirSLAM follows a two-stage offline approach. First, potential candidates are retrieved using DBoW2. In the loop correction stage, PGO is used to correct the estimated trajectory, followed by global BA to ensure global consistency.

Since our testing conditions differ from the default examples provided by these methods, careful parameter tuning was essential. To ensure robust performance in our unique environment, we empirically tuned the feature extraction parameters within the frontend of each SLAM algorithm.

### 3.3 Evaluation Metrics and Criteria

In Visual and Visual-Inertial SLAM systems, maintaining the global accuracy of the predicted trajectory is crucial. This accuracy is evaluated by measuring the absolute and relative differences between the estimated trajectory and the ground truth trajectory. Since these trajectories may be presented in different coordinate systems, they must be aligned first. Umeyama's method (Umeyama, 1991) can be used to solve this alignment in closed form. The method identifies the transformation that yields the optimal least-squares solution to map the estimated trajectory onto the ground truth trajectory.

We compute the root mean squared (RMSE) absolute trajectory error (ATE) as one of the main metrics to quantify the deviation between estimated and ground truth positions. The ATE measures the global consistency of the estimated trajectory, reflecting the overall difference between the estimated and ground truth positions after alignment. In addition to ATE, we compute the RMSE relative pose error (RPE) to evaluate the local accuracy of the SLAM algorithms over short intervals. The RPE measures the error in the relative transformations between consecutive poses, thus evaluating the local accuracy and consistency of the trajectory. For the calculation of both ATE and RPE, we follow the methodologies described in (Sturm et al., 2012). We further analyze CPU and memory utilization to understand the computational demands of each Visual-Inertial SLAM method and LC. Since we rely on ROS-based algorithms, the CPU and memory utilization of the corresponding ROS nodes can be determined accurately.

To verify the reliability and reproducibility of our results, we evaluate the SLAM algorithms by conducting five trials on each dataset sequence before calculating the average RMSE ATE and RPE using the open-source evaluation toolbox Evo (Grupp, 2017). We filter out invalid trajectories to ensure the accuracy and reliability of the analysis by excluding failed runs. The criteria for filtering invalid trajectories include:

- **Scenario Duration Coverage:** Runs with valid pose estimates in less than 80% of the scenario duration are considered failures. Thus, only trajectories with sufficient coverage and consistency are analyzed.
- **Pose Frequency:** Runs must produce at least one pose per second to be deemed successful, ensuring adequate temporal resolution for precise tracking.

Trajectories failing to meet these criteria are excluded from further analysis. This process is essential for maintaining data integrity and focusing on runs that adequately reflect the SLAM systems' performance. We conducted all experiments on an Intel Core i9-13900HX with 64 GB RAM, operating within a Docker container based on Ubuntu 20.04 and ROS Noetic.

## 4 Experimental Evaluations

In this section, we present the results from the experiments. We investigate the influence of implementing LC on the localization accuracy and computational demands of various SLAM algorithms. ORB-SLAM3 was evaluated in both mono-inertial (M-I) and stereo-inertial (M-I) configurations, HFNet-SLAM in mono-inertial, and all other methods in stereo-inertial mode, operating on the data provided by the Intel RealSense T265.

### 4.1 Localization Accuracy

To evaluate the localization accuracy of the SLAM algorithms, we focus on both quantitative metrics and qualitative observations.

### 4.1.1 Quantitative Results

We use the ATE and RPE metrics to evaluate global and local localization accuracy, respectively. Tables 3 and 4 present the RMSE ATE results for mono-inertial and stereo-inertial configurations, respectively, while Tables 5 and 6 provide the RMSE RPE results for mono-inertial and stereo-inertial setups across the different scenarios.

**ORB-SLAM3 (M-I)** showed variable performance across environments, with frame rate and LC significantly affecting results. In the *Garden Medium* environment, the method struggled considerably, particularly in the *Perimeter* scenario, where performance was very poor at lower frame rates and without LC. The *Lane* scenario failed entirely. Performance in the *Garden Large* environment was similarly poor, with the *Perimeter* scenario failing to work at all. The *Lane* scenario also showed very poor results, particularly at lower frame rates and without LC. In the *Park* environment, the *Perimeter* scenario's performance was significantly impacted by frame rate and LC, with optimal results at 30 FPS and with LC enabled. The *Lane* scenario demonstrated stable performance, with higher frame rates yielding comparable results and LC providing a performance boost, particularly at 30 FPS. RPE was moderate to poor, with LC improving results in the *Perimeter* scenario. Overall, ORB-SLAM3 (M-I) exhibited moderate performance in the *Garden Small* and *Park* environments but significantly struggled in the *Garden Medium* and *Garden Large* environments. Higher frame rates led to the best results, with 5 FPS showing heavy struggles. The application of LC notably improved performance, particularly in the *Perimeter* scenario.

**HFNet-SLAM** performed best at lower frame rates, particularly in the *Perimeter* scenario, where LC improved accuracy, while *Lane* showed little impact from LC and remained more stable across configurations, demonstrating consistent performance across locations. In the *Garden Small* environment, the *Perimeter* scenario achieved lower ATE at reduced frame rates, with LC having no significant impact. The *Lane* scenario performed best at 30 FPS with LC, while lower frame rates showed minimal variation. In the *Garden Medium* environment, the *Lane* scenario failed consistently, while the *Perimeter* scenario performed best at 15 FPS, with LC improving results across all frame rates. In the *Garden Large* environment, the *Perimeter* scenario benefited from lower frame rates, though LC led to fluctuating results. The *Lane* scenario performed best at higher frame rates but showed inconsistencies with LC. In the *Park* environment, the *Perimeter* scenario achieved the best results at lower frame rates with LC, while the *Lane* scenario remained stable across all configurations. HFNet-SLAM showed moderate RPE in the *Perimeter* scenario, with lower frame rates and LC yielding the best results. In the *Lane* scenario, RPE remained consistently low across all configurations, with optimal performance at 15 and 30 FPS and no significant impact from LC. Overall, HFNet-SLAM achieved moderate accuracy and benefited from lower frame rates, with LC improving *Perimeter* but having little to no effect on *Lane* results.

**ORB-SLAM3 (S-I)** exhibited consistent performance with and without LC across different frame rates in multiple environments, particularly maintaining low ATE values for both *Perimeter* and *Lane* scenarios. In the *Garden Small* environment, the *Perimeter* scenario presented stable results for the different configurations, with only a minor increase of ATE in lower frame rates. In contrast, the performance of the *Lane* scenario significantly declined at 5 FPS, although a very low ATE was achieved at higher frame rates. In the *Garden Medium* environment, challenges were more pronounced, especially for the *Lane* scenario, which failed consistently. The *Perimeter* scenario performed better with LC and higher frame rates. In the *Garden Large* environment, the method faces

**TABLE 3** RMSE ATE [m] of mono-inertial SLAM methods across different environments with and without LC at 30 FPS, 15 FPS, and 5 FPS for the *Perimeter* (P) and *Lane* (L) scenario. An 'x' indicates that the method failed, either due to no valid results being available or errors exceeding 100 meters, implying significant drift and unreliable position estimation in the tested environments. To calculate the scenario-specific mean average, values were only considered when results were reported across all FPS and LC configurations of the method. The best method-specific result is in bold, with the top three results across all methods highlighted as best, second, and third.

| | | | Garden Small | | Garden Medium | | Garden Large | | Park | | Average | |
|---|---|---|---|---|---|---|---|---|---|---|---|---|
| **Method** | **FPS** | **LC** | P | L | P | L | P | L | P | L | P | L |
| ORB-SLAM3 | 30 | On | **0.82** | 0.26 | **5.59** | x | x | 8.87 | **1.91** | **0.26** | **1.91** | **0.26** |
| | | Off | 1.96 | 0.27 | 39.73 | x | x | 81.64 | 2.39 | 0.31 | 2.39 | 0.29 |
| | 15 | On | 1.55 | **0.23** | 16.44 | x | x | **7.80** | 2.00 | 0.30 | 2.00 | 0.27 |
| | | Off | x | 0.27 | 58.19 | x | x | x | 2.09 | 0.38 | 2.09 | 0.33 |
| | 5 | On | 2.55 | 0.31 | x | x | x | x | 28.22 | 0.86 | 28.22 | 0.58 |
| | | Off | 56.70 | 0.34 | x | x | x | x | 88.58 | 8.37 | 88.58 | 4.35 |
| HFNet-SLAM | 30 | On | 1.14 | **0.26** | 1.36 | x | 2.51 | 0.98 | 1.25 | 0.33 | 1.57 | 0.52 |
| | | Off | 0.99 | 0.32 | 1.82 | x | 2.30 | **0.76** | 1.23 | **0.26** | 1.58 | **0.45** |
| | 15 | On | 0.97 | 0.32 | **1.05** | x | 2.18 | 0.81 | **0.94** | 0.27 | **1.28** | 0.47 |
| | | Off | 1.02 | 0.32 | 1.80 | x | **1.97** | 1.05 | 0.99 | 0.34 | 1.58 | **0.45** |
| | 5 | On | **0.95** | 0.31 | 1.24 | x | **1.97** | 1.05 | 0.99 | 0.34 | 1.29 | 0.56 |
| | | Off | **0.95** | 0.31 | 1.89 | x | 2.10 | 0.95 | 1.74 | 0.29 | 1.67 | 0.52 |

**TABLE 4** RMSE ATE [m] of stereo-inertial SLAM methods across different environments with and without LC at 30 FPS, 15 FPS, and 5 FPS for the *Perimeter* (P) and *Lane* (L) scenario (cf. Table 3 for details on failure cases and calculation of the average).

| Method | FPS | LC | Garden Small | | Garden Medium | | Garden Large | | Park | | Average | |
|---|---|---|---|---|---|---|---|---|---|---|---|---|
| | | | P | L | P | L | P | L | P | L | P | L |
| ORB-SLAM3 | 30 | On | **0.24** | 0.19 | 0.64 | x | **1.68** | **0.36** | **0.65** | **0.26** | **0.51** | 0.19 |
| | | Off | **0.24** | **0.18** | 0.75 | x | x | 0.80 | 0.68 | 0.30 | 0.56 | **0.18** |
| | 15 | On | 0.25 | 0.19 | **0.60** | x | 1.69 | **0.36** | 0.67 | 0.33 | **0.51** | 0.19 |
| | | Off | 0.28 | **0.18** | 0.73 | x | x | 1.62 | 0.70 | 0.38 | 0.57 | **0.18** |
| | 5 | On | 0.26 | 0.50 | 0.85 | x | x | x | 2.58 | x | 1.23 | 0.50 |
| | | Off | 0.30 | 0.61 | 1.45 | x | 1.89 | x | 3.27 | x | 1.67 | 0.61 |
| VINS-Fusion | 30 | On | 1.64 | x | 2.02 | **0.58** | 3.90 | x | **0.79** | **20.70** | **1.41** | **0.58** |
| | | Off | 1.69 | **0.21** | 2.08 | 0.59 | 4.00 | x | 1.81 | 24.86 | 1.94 | 0.59 |
| | 15 | On | **1.47** | x | **1.94** | 1.36 | **3.86** | **19.65** | 1.63 | 92.37 | 1.79 | 1.36 |
| | | Off | 1.51 | x | 2.00 | 1.37 | 3.89 | 19.81 | 2.74 | 89.40 | 2.37 | 1.37 |
| | 5 | On | x | x | 2.43 | 2.64 | x | x | 15.89 | x | 9.16 | 2.64 |
| | | Off | x | x | 2.87 | 2.89 | x | x | 16.69 | x | 9.78 | 2.89 |
| OpenVINS + VINS-Fusion + HLoc | 30 | VF | 0.50 | x | 0.82 | **0.35** | **0.75** | x | 1.58 | **0.19** | **0.91** | **0.19** |
| | | HLoc | 0.71 | 0.35 | 0.83 | 0.47 | 0.85 | **3.04** | **1.52** | 0.68 | 0.98 | 0.68 |
| | | Off | 0.52 | x | 1.02 | 0.41 | 0.95 | x | 1.67 | 0.31 | 1.04 | 0.31 |
| | 15 | VF | 0.49 | x | 0.82 | x | **0.75** | x | 1.58 | **0.19** | **0.91** | **0.19** |
| | | HLoc | 0.71 | 0.35 | 0.84 | 0.42 | 0.89 | x | 1.51 | 0.68 | 0.99 | 0.68 |
| | | Off | 0.51 | x | 1.02 | x | 0.95 | x | 1.67 | 0.31 | 1.04 | 0.31 |
| | 5 | VF | **0.45** | x | **0.64** | x | 1.46 | x | 1.80 | 0.33 | 1.09 | 0.33 |
| | | HLoc | 0.69 | **0.34** | 0.93 | 0.91 | 1.85 | 22.28 | 2.23 | 0.53 | 1.42 | 0.53 |
| | | Off | 0.58 | x | 0.86 | x | 1.81 | x | 1.93 | 0.36 | 1.29 | 0.36 |
| OpenVINS + Maplab | 30 | On | 0.32 | 0.20 | **0.43** | **0.22** | **0.65** | 5.63 | **0.56** | **0.17** | **0.49** | **0.20** |
| | | Off | 0.61 | 0.26 | 1.24 | 0.40 | 1.06 | **4.93** | 1.63 | **0.17** | 1.13 | 0.28 |
| | 15 | On | **0.27** | **0.18** | 0.44 | 0.23 | 0.66 | 8.19 | 0.76 | 0.19 | 0.53 | **0.20** |
| | | Off | 0.59 | 0.26 | 1.24 | 0.40 | 1.06 | 6.63 | 1.63 | 0.19 | 1.13 | 0.28 |
| | 5 | On | 0.71 | 0.81 | 0.91 | 0.84 | 1.56 | x | 1.76 | 0.81 | 1.24 | 0.82 |
| | | Off | 0.68 | 0.19 | 0.98 | 0.87 | 2.03 | x | 2.17 | 0.33 | 1.47 | 0.46 |
| Kimera | 30 | On | 1.69 | 7.82 | 3.50 | 91.41 | 11.67 | x | x | 50.14 | 5.62 | 50.14 |
| | | Off | 1.02 | x | 1.90 | x | 6.75 | x | 21.89 | 31.87 | 3.22 | 31.87 |
| | 15 | On | 2.75 | x | 2.76 | **6.81** | 9.58 | x | 22.91 | 17.53 | 5.03 | 17.53 |
| | | Off | **0.99** | **0.47** | **1.72** | 22.30 | **5.28** | x | **4.34** | 31.39 | **2.66** | 31.39 |
| | 5 | On | 6.87 | x | 6.24 | x | 16.86 | x | x | 45.79 | 9.99 | 45.79 |
| | | Off | 6.09 | 9.55 | 3.96 | x | 29.99 | x | 62.99 | **2.96** | 13.35 | **2.96** |
| SVO Pro | 30 | On | **1.15** | **0.29** | **1.48** | **1.14** | **3.49** | x | **2.14** | **0.61** | **2.49** | x |
| | | Off | 1.29 | 0.31 | 1.59 | x | 4.18 | x | 2.32 | 0.69 | 2.88 | x |
| | 15 | On | 1.56 | 0.39 | 2.06 | x | 4.33 | x | 2.83 | 1.01 | 3.19 | x |
| | | Off | 1.80 | 0.96 | 2.06 | x | 5.23 | x | 2.71 | 1.14 | 3.64 | x |
| | 5 | On | x | x | 8.66 | x | 58.82 | x | x | x | 33.74 | x |
| | | Off | x | x | 34.37 | x | 51.00 | x | x | x | 42.68 | x |
| AirSLAM | 30 | On | 1.60 | 2.77 | 9.32 | 4.56 | 15.29 | 5.53 | 6.31 | 7.67 | 8.13 | 5.13 |
| | | Off | 4.74 | 3.18 | 9.33 | 4.56 | 15.39 | 5.53 | 6.24 | 8.21 | 8.92 | 5.37 |
| | 15 | On | **1.48** | 2.43 | 8.92 | 4.14 | 16.68 | 5.00 | **5.05** | **2.56** | 8.03 | 3.53 |
| | | Off | 4.39 | 2.83 | 8.91 | 4.14 | 16.68 | 5.00 | 6.46 | 2.64 | 9.11 | 3.65 |
| | 5 | On | 5.82 | 1.73 | **5.31** | **0.74** | **7.36** | **2.90** | 6.86 | 5.73 | **6.34** | **2.78** |
| | | Off | 5.78 | **1.67** | **5.31** | 1.83 | 10.81 | 3.80 | 6.98 | 5.70 | 7.22 | 3.25 |

challenges in the *Perimeter* scenario with high ATE values and partial failures. However, LC significantly improves ATE in the *Lane* scenario at higher frame rates, though it fails entirely at lower frame rates. The *Park* environment showed better performance with higher frame rates, with LC providing some improvement, although challenges persisted at lower frame rates. RPE remained low with LC across all environments but increased at lower frame rates and without LC, particularly in more complex environments such as *Garden Large* and *Park*. Overall, ORB-SLAM3 (S-I) benefits significantly from LC and higher frame rates across all tested environments, with LC providing a notable decrease in ATE for both *Lane* and *Perimeter* scenarios.

**VINS-Fusion** showed similar performance across multiple environments. LC consistently improved results in both *Perimeter* and *Lane* scenarios, particularly at higher frame rates. In the *Garden Small*, the *Perimeter* scenario performed better at 15 FPS compared to 30 FPS, with LC slightly reducing the ATE. The *Lane* scenario struggled across all frame rates, with significant challenges at lower frame rates. In the *Garden Medium* environment, LC reduced ATE for both scenarios. While lower frame rates increased ATE in the *Perimeter* scenario, the performance at 30 FPS and 15 FPS remained comparable. The *Lane* scenario's ATE increased significantly as the frame rate decreased. In the *Garden Large*, LC lowered ATE for the *Perimeter* scenario, showing comparable performance at 30 FPS and 15 FPS. However, the algorithm failed for the *Lane* scenario at 30 FPS and for both scenarios at 5 FPS. In the *Park* environment, LC improved results for the *Perimeter* scenario, but ATE increased significantly as the frame rate decreased. The *Lane* scenario consistently showed poor performance, failing at lower frame rates. RPE results indicated good local accuracy with LC, especially at higher frame rates. Lower frame rates led to higher RPE values, in particular in the *Garden Small*, *Garden Medium*, and *Park* environments. Overall, VINS-Fusion performed best at 30 FPS with LC, which reduced both ATE and RPE. However, the algorithm encountered difficulties at lower frame rates and in the *Lane* scenario across different environments.

**OpenVINS+VINS-Fusion** achieved better accuracy than **OpenVINS+HLoc** in the *Garden Small* environment for the *Perimeter* scenario, while OpenVINS without LC exhibited the highest error. The frame rate had little impact on OpenVINS+HLoc, maintaining stable but lower accuracy than OpenVINS+VINS-Fusion. However, in the *Lane* scenario, both OpenVINS+VINS-Fusion and without LC failed to produce valid results, whereas OpenVINS+HLoc provided stable outputs across all frame rates, demonstrating improved robustness in this setting. In the *Garden Medium* environment, a similar trend emerged. For *Perimeter*, OpenVINS+VINS-Fusion performed slightly better than OpenVINS+HLoc, with without LC yielding the highest error. However, OpenVINS+HLoc remained more resilient to frame rate reductions, continuing to provide valid results where OpenVINS+VINS-Fusion and without LC failed in the *Lane* scenario. Lowering the frame rate degraded OpenVINS+HLoc performance, but it still outperformed the other configurations in terms of robustness. In Garden Large, the frame rate had a significant effect on all configurations. For the *Perimeter* scenario, OpenVINS+VINS-Fusion achieved the lowest ATE, followed by OpenVINS+HLoc, while without LC, it produced the highest error. Performance degradation with decreasing frame rate was most pronounced for OpenVINS+HLoc, suggesting a stronger dependency on frequent updates. In the *Lane* scenario, all methods struggled, with OpenVINS+HLoc also failing to provide reliable results. In the *Park* environment, OpenVINS+HLoc achieved slightly better results than OpenVINS+VINS-Fusion and without LC in the *Perimeter* scenario at higher frame rates, but at 5 FPS, its performance declined sharply. In contrast, OpenVINS+VINS-Fusion remained more stable across frame rates. However, in the *Lane* scenario, OpenVINS+HLoc performed significantly worse than both OpenVINS+VINS-Fusion and without LC, suggesting that its robustness in *Lane* is location-dependent and not consistent across all environments. For RPE, OpenVINS+VINS-Fusion exhibited strong local accuracy across all frame rates in *Perimeter*, with increasing RPE at lower frame rates, especially in *Lane*. OpenVINS+HLoc, however, showed higher RPE across all frame rates and scenarios compared to both OpenVINS+VINS-Fusion and without LC, suggesting weaker local consistency despite its robustness in some *Lane* scenarios. Overall, OpenVINS+VINS-Fusion delivered the most consistent performance, particularly in *Perimeter* scenarios and across different environments. OpenVINS+HLoc performed worse in *Perimeter* than OpenVINS+VINS-Fusion but better than without LC at higher frame rates. Its major advantage was its robustness in *Lane* scenarios, where it produced valid results while OpenVINS+VINS-Fusion and without LC often failed. However, this robustness did not generalize well

**TABLE 5** RMSE RPE [m] per meter of mono-inertial SLAM methods across different environments with and without LC at 30 FPS, 15 FPS, and 5 FPS in the *Perimeter* (P) and *Lane* (L) scenario. An 'x' indicates that the method failed, either due to no valid results being available or errors exceeding 1 meter, implying significant drift and unreliable position estimation in the tested environments. To calculate the scenario-specific mean average, values were only considered when results were reported across all FPS and LC configurations of the method.

| | | | Garden Small | | Garden Medium | | Garden Large | | Park | | Average | |
|---|---|---|---|---|---|---|---|---|---|---|---|---|
| **Method** | **FPS** | **LC** | P | L | P | L | P | L | P | L | P | L |
| ORB-SLAM3 | 30 | On | **0.23** | **0.17** | **0.30** | x | 0.44 | **0.30** | **0.30** | **0.15** | **0.32** | **0.21** |
| | | Off | 0.25 | 0.18 | 0.39 | x | 0.48 | 0.37 | 0.31 | 0.16 | 0.35 | 0.23 |
| | 15 | On | 0.27 | 0.18 | 0.46 | x | **0.37** | 0.35 | 0.34 | **0.15** | **0.32** | 0.23 |
| | | Off | 0.35 | **0.17** | 0.47 | x | 0.45 | 0.48 | 0.37 | 0.17 | 0.39 | 0.27 |
| | 5 | On | 0.35 | 0.21 | x | x | 0.44 | 0.43 | 0.38 | 0.39 | 0.39 | 0.34 |
| | | Off | 0.43 | 0.23 | x | x | 0.57 | 0.46 | 0.41 | 0.49 | 0.47 | 0.39 |
| HFNet-SLAM | 30 | On | 0.28 | 0.20 | 0.21 | x | 0.26 | 0.17 | 0.19 | 0.17 | 0.24 | 0.18 |
| | | Off | **0.25** | **0.19** | 0.24 | x | 0.23 | **0.15** | 0.20 | 0.17 | 0.23 | **0.17** |
| | 15 | On | **0.25** | **0.19** | **0.17** | x | 0.22 | 0.16 | **0.16** | 0.17 | **0.20** | **0.17** |
| | | Off | **0.25** | **0.19** | 0.24 | x | 0.23 | **0.15** | 0.19 | **0.16** | 0.23 | **0.17** |
| | 5 | On | 0.26 | **0.19** | 0.21 | x | 0.22 | 0.25 | 0.17 | 0.18 | 0.21 | 0.21 |
| | | Off | 0.26 | **0.19** | 0.25 | x | **0.21** | 0.23 | 0.25 | 0.17 | 0.24 | 0.20 |

**TABLE 6** RMSE RPE [m] per meter of stereo-inertial SLAM methods across different environments with and without LC at 30 FPS, 15 FPS, and 5 FPS in the *Perimeter* (P) and *Lane* (L) scenario (cf. Table 5 for details on failure cases and calculation of the average).

| | | | Garden Small | | Garden Medium | | Garden Large | | Park | | Average | |
|---|---|---|---|---|---|---|---|---|---|---|---|---|
| **Method** | **FPS** | **LC** | P | L | P | L | P | L | P | L | P | L |
| ORB-SLAM3 | 30 | On | **0.09** | **0.16** | **0.07** | x | **0.10** | **0.10** | **0.09** | **0.13** | **0.09** | **0.16** |
| | | Off | **0.09** | 0.17 | 0.08 | x | 0.16 | 0.12 | **0.09** | 0.15 | 0.10 | 0.17 |
| | 15 | On | **0.09** | 0.17 | **0.07** | x | 0.12 | **0.10** | 0.10 | 0.15 | 0.10 | 0.17 |
| | | Off | **0.09** | 0.16 | **0.07** | x | 0.15 | 0.14 | 0.11 | 0.18 | 0.10 | **0.16** |
| | 5 | On | **0.09** | 0.17 | 0.10 | x | 0.25 | x | 0.27 | x | 0.18 | 0.17 |
| | | Off | **0.09** | 0.18 | 0.12 | x | 0.19 | 0.23 | 0.35 | x | 0.19 | 0.18 |
| VINS-Fusion | 30 | On | 0.17 | x | 0.14 | **0.11** | **0.13** | 0.33 | **0.14** | 0.32 | **0.14** | 0.22 |
| | | Off | 0.18 | **0.17** | **0.13** | 0.12 | **0.13** | 0.36 | **0.14** | **0.31** | 0.15 | 0.24 |
| | 15 | On | **0.13** | x | **0.13** | 0.16 | **0.13** | 0.21 | 0.18 | 0.39 | **0.14** | 0.19 |
| | | Off | **0.13** | 0.25 | **0.13** | 0.16 | **0.13** | **0.20** | 0.18 | 0.38 | **0.14** | **0.18** |
| | 5 | On | 0.78 | x | 0.23 | 0.23 | 0.53 | 0.21 | 0.36 | x | 0.48 | 0.22 |
| | | Off | 0.18 | 0.30 | 0.25 | 0.25 | 0.31 | **0.20** | 0.34 | 0.96 | 0.27 | 0.22 |
| OpenVINS + VINS-Fusion + HLoc | 30 | VF | 0.13 | x | 0.13 | **0.11** | **0.08** | 0.20 | 0.25 | 0.12 | 0.15 | **0.14** |
| | | HLoc | 0.22 | 0.23 | 0.17 | 0.15 | 0.15 | 0.21 | 0.27 | 0.23 | 0.20 | 0.20 |
| | | Off | 0.13 | x | 0.12 | 0.15 | 0.13 | 0.24 | 0.24 | 0.11 | 0.15 | 0.17 |
| | 15 | VF | 0.12 | x | 0.13 | 0.24 | **0.08** | **0.19** | 0.25 | **0.10** | 0.15 | 0.18 |
| | | HLoc | 0.22 | 0.23 | 0.17 | 0.15 | 0.15 | 0.25 | 0.27 | 0.24 | 0.20 | 0.22 |
| | | Off | 0.14 | x | 0.12 | 0.23 | 0.13 | 0.24 | 0.24 | 0.14 | 0.16 | 0.20 |
| | 5 | VF | 0.13 | x | **0.11** | 0.31 | 0.13 | 0.24 | **0.20** | 0.17 | **0.14** | 0.24 |
| | | HLoc | 0.22 | 0.23 | 0.20 | 0.24 | 0.17 | 0.31 | 0.24 | 0.25 | 0.21 | 0.27 |
| | | Off | 0.12 | x | **0.11** | 0.36 | 0.13 | 0.23 | 0.23 | 0.23 | 0.15 | 0.27 |
| OpenVINS + Maplab | 30 | On | **0.11** | 0.13 | **0.10** | **0.10** | 0.10 | **0.19** | 0.15 | 0.14 | **0.12** | **0.12** |
| | | Off | 0.13 | 0.15 | 0.13 | **0.10** | **0.09** | 0.20 | 0.23 | **0.10** | 0.15 | **0.12** |
| | 15 | On | **0.11** | **0.12** | 0.11 | **0.10** | 0.11 | 0.21 | 0.16 | 0.15 | **0.12** | **0.12** |
| | | Off | 0.13 | 0.15 | 0.13 | **0.10** | **0.09** | 0.23 | 0.23 | **0.10** | 0.14 | **0.12** |
| | 5 | On | 0.14 | 0.26 | **0.10** | 0.17 | 0.12 | 0.32 | **0.14** | 0.62 | **0.12** | 0.35 |
| | | Off | 0.13 | 0.15 | 0.13 | 0.22 | 0.14 | x | 0.22 | 0.16 | 0.16 | 0.17 |
| Kimera | 30 | On | 0.18 | x | 0.18 | **0.25** | 0.22 | 0.33 | 0.53 | 0.81 | 0.28 | 0.46 |
| | | Off | **0.16** | **0.21** | **0.16** | **0.25** | 0.20 | **0.29** | 0.41 | **0.64** | **0.23** | **0.39** |
| | 15 | On | 0.21 | x | 0.21 | 0.36 | 0.21 | **0.29** | 0.44 | 0.69 | 0.27 | 0.45 |
| | | Off | **0.16** | **0.21** | 0.18 | 0.32 | **0.19** | 0.39 | **0.38** | 0.69 | **0.23** | 0.47 |
| | 5 | On | 0.27 | x | 0.28 | 0.46 | 0.34 | 0.55 | 0.88 | 0.79 | 0.44 | 0.60 |
| | | Off | 0.24 | 0.32 | 0.27 | 0.41 | 0.43 | 0.66 | 0.83 | 0.61 | 0.44 | 0.50 |
| SVO Pro | 30 | On | **0.14** | **0.16** | **0.14** | 0.34 | **0.16** | **0.26** | **0.17** | **0.14** | **0.15** | 0.22 |
| | | Off | **0.14** | **0.16** | 0.15 | **0.20** | **0.16** | 0.28 | **0.17** | **0.14** | **0.15** | **0.19** |
| | 15 | On | 0.17 | 0.17 | 0.16 | 0.73 | 0.19 | **0.26** | 0.20 | 0.19 | 0.18 | 0.34 |
| | | Off | 0.17 | 0.18 | 0.15 | 0.63 | 0.20 | 0.32 | 0.20 | 0.19 | 0.18 | 0.33 |
| | 5 | On | 0.40 | 0.40 | 0.39 | 0.95 | 0.41 | 0.38 | 0.38 | 0.45 | 0.39 | 0.54 |
| | | Off | 0.40 | 0.36 | 0.41 | 0.95 | 0.42 | 0.37 | 0.42 | 0.49 | 0.41 | 0.54 |
| AirSLAM | 30 | On | 0.19 | 0.34 | 0.13 | 0.18 | **0.20** | **0.19** | **0.21** | 0.26 | **0.18** | **0.24** |
| | | Off | **0.18** | 0.39 | 0.13 | 0.19 | **0.20** | 0.20 | **0.21** | 0.28 | **0.18** | 0.27 |
| | 15 | On | 0.21 | 0.30 | **0.12** | **0.16** | 0.28 | 0.49 | 0.22 | **0.16** | 0.21 | 0.28 |
| | | Off | 0.21 | 0.29 | **0.12** | 0.17 | 0.29 | 0.64 | 0.23 | 0.17 | 0.21 | 0.32 |
| | 5 | On | 0.25 | 0.26 | 0.15 | 0.18 | 0.48 | 0.49 | 0.22 | 0.26 | 0.27 | 0.30 |
| | | Off | 0.25 | **0.25** | 0.14 | 0.18 | 0.56 | 0.95 | 0.22 | 0.30 | 0.29 | 0.42 |

to larger environments or lower frame rates, where its performance deteriorated significantly.

**OpenVINS+Maplab** demonstrated better performance with LC for multiple environments at higher frame rates. In the *Garden Small* environment, LC effectively reduced ATE for both *Perimeter* and *Lane* scenarios at 30 FPS and 15 FPS. In the *Garden Medium* environment, performance declined notably at 5 FPS, especially for the *Lane* scenario. In the *Garden Large* environment, LC reduced ATE for the *Perimeter* scenario at 30 FPS and 15 FPS, but the *Lane* scenario struggled, particularly at lower frame rates, where it failed completely. In the *Park* environment, the *Perimeter* scenario consistently showed lower ATE with LC at higher frame rates, while the *Lane* scenario's ATE was not significantly impacted by LC and increased at 5 FPS. RPE results indicated that LC helped maintain low values at high frame rates, with higher RPE values without LC, and at lower frame rates in complex environments. Overall, OpenVINS+Maplab performed well with LC, particularly at higher frame rates, but it struggled in complex scenarios and at lower frame rates.

**Kimera's** performance showed consistently high ATE values across the different environments, particularly struggling in the *Lane* scenario. In the *Garden Small* environment, ATE increased with LC and worsened at lower frame rates. In the *Garden Medium* and *Garden Large* environments, ATE increased significantly and often failed for the *Lane* scenarios. The *Park* environment saw irregular performance across frame rates with very high ATE values. RPE results also highlighted local accuracy challenges, with RPE values higher with LC across most scenarios and significantly increasing at 5 FPS. Overall, Kimera struggled with both ATE and RPE in the *Lane* scenarios and larger environments, with LC often worsening these issues.

**SVO Pro** showed improved performance with LC in various environments, particularly at higher frame rates. For the *Garden Small* environment, LC effectively lowered ATE for both *Perimeter* and *Lane* scenarios, notably at 15 FPS. The *Garden Medium* environment had a lower ATE for the *Perimeter* scenario with LC at 30 FPS and 5 FPS, whereas the *Lane* scenario struggled significantly. In the *Garden Large* environment, LC improved ATE for the *Perimeter* scenario, with lower ATE at higher frame rates, while the *Lane* scenario failed completely. In the *Park* environment, ATE was reduced with LC at 30 FPS, but performance degraded as frame rates decreased, with both scenarios failing at 5 FPS. RPE results indicated reasonable local accuracy with LC, increasing slightly without LC and significantly at lower frame rates, particularly in the *Garden Medium* and *Park*. Overall, SVO Pro benefits from LC and higher frame rates but struggles at lower frame rates and in complex environments.

**AirSLAM** exhibited overall poor localization accuracy across both driving scenarios, with only slight improvements when LC was enabled. ATE performance consistently deteriorated as the environment size increased, suggesting that the method struggles with large-scale mapping and drift correction. RPE results were moderate to poor, particularly in the *Lane* scenario, indicating challenges in maintaining local trajectory consistency. Interestingly, ATE improved at lower frame rates, while RPE worsened, indicating that reducing the update frequency helped mitigate global drift but negatively affected short-term pose estimation accuracy. These results highlight AirSLAM's difficulty in maintaining both global and local consistency, particularly in larger environments and at higher frame rates.

### 4.1.2 Qualitative Results

We qualitatively evaluate the SLAM algorithms by comparing the trajectories they generated across all environments. Specifically, we evaluate HFNet-SLAM at 15 Hz with LC, along with ORB-SLAM3 in stereo-inertial setup, VINS-Fusion, OpenVINS+VINS-Fusion, OpenVINS+Maplab, SVO Pro, and Kimera at 30 FPS with LC against the ground truth, see Figures 2 to 6 (a). Additionally, we present the trajectories of a selected method under various frame rates and LC configurations, as illustrated in Figures 2 through 6 (b). For the sake of clarity, we did not include ORB-SLAM3 in the mono-inertial setup, OpenVINS+HLoc, and AirSLAM in the qualitative evaluation.

In the *Garden Small Perimeter* scenario, ORB-SLAM3, OpenVINS with Maplab, and OpenVINS with VINS-Fusion demonstrated trajectories close to the ground truth, showing good resistance to drift over time (see Figure 2a). While HFNet-SLAM also showed good resistance to drift, it faced scaling issues. Conversely, SVO Pro, VINS-Fusion, and Kimera experienced significant drift. When comparing different LC and FPS configurations of OpenVINS+Maplab, we observed that at 30 FPS and 15 FPS, LC helped to produce highly accurate results, while without LC, the algorithms struggled with scale, see Figure 2b. At 5 FPS, significant issues arose in both scenarios, particularly during rotations.

For the *Garden Medium Perimeter* scenario, Maplab generated very accurate trajectories, whereas ORB-SLAM3, OpenVINS+VINS-Fusion, and HFNet-SLAM were accurate but had minor scale issues. SVO Pro and VINS-Fusion struggled with scale and minor drift, and Kimera encountered substantial drift, as visualized in Figure 3a. Across different frame rates and LC configurations of ORB-SLAM3, we observed minor drift without LC at 30 FPS and 15 FPS. We further observed that drift increased significantly at 5 FPS, especially without LC, emphasizing the importance of LC at lower frame rates, see Figure 3b.

In the *Garden Large Perimeter* scenario, OpenVINS+Maplab and OpenVINS+VINS-Fusion maintained accurate trajectories with minor scale issues, depicted in Figure 4a. ORB-SLAM3 and HFNet-SLAM experienced scale issues and minor drift, while VINS-Fusion and SVO Pro encountered significant drift. Kimera failed completely in this environment. Comparatively, with OpenVINS+Maplab at 30 FPS and 15 FPS with LC, the results were consistent with minor scaling issues, but without LC, drift became more apparent. At 5 FPS, trajectories showed significant scaling and drift problems, which worsened without LC (see Figure 4b).

In the *Park Perimeter* scenario, ORB-SLAM3, OpenVINS+Maplab, and VINS-Fusion performed well with minor scale issues. OpenVINS+VINS-Fusion and HFNet-SLAM had some scale issues, while SVO Pro experienced significant drift and Kimera failed completely, as visualized in Figure 5a. When comparing different frame rates and LC configurations, see Figure 5b, ORB-SLAM3 showed comparable results at 30 FPS and 15 FPS, both with and without LC, showing minor scale issues. At 5 FPS, the system exhibited pronounced scale inconsistencies and considerable drift when operating without LC.

In the *Garden Medium Lane* scenario, OpenVINS+Maplab and OpenVINS+VINS-Fusion were accurate with minor scale issues and managed 180° rotations effectively, as shown in Figure 6a. VINS-Fusion showed drift during 180° rotations and had scale issues, while SVO Pro had larger scale issues and drifted away toward the end. HFNet-SLAM, ORB-SLAM3, and Kimera failed completely. For OpenVINS+Maplab, at 30 FPS and 15 FPS without LC, the results were nearly identical, struggling with scale and drift. With LC, the results improved, showing minor scale issues. At 5 FPS, both with and without LC, we observed significant scale problems (see Figure 6b).

## 4.2 Computational Efficiency

The implementation of LC significantly impacts the computational resources required by SLAM algorithms. We provide detailed quantitative evaluations for the presented SLAM algorithms of a specific run in the *Garden Medium Perimeter* scenario in Table 7. OpenVINS+Maplab and AirSLAM are not included, as they operate offline in a post-processing manner.

Experimental evaluations reveal that **ORB-SLAM3** exhibits high CPU usage compared to other evaluated SLAM methods, with the mono-inertial configuration consuming slightly more resources than the stereo-inertial setup. Memory usage remains moderate in both cases, with minimal differences between LC-enabled and LC-disabled modes. The stereo-inertial setup maintains stable memory consumption at 30 and 15 FPS, with a slight decrease at lower frame rates, whereas the mono-inertial setup requires roughly 50% more memory than its stereo counterpart. This resource utilization pattern can be attributed to ORB-SLAM3's design: ORB features are extracted directly in the frontend and seamlessly integrated into the DBoW2 loop detection module, minimizing additional overhead when LC is enabled. The high CPU load is primarily driven by the computationally intensive local BA and the real-time feature extraction and descriptor matching that add to the processing cost, particularly at higher frame rates. However, since PGO and global BA are triggered only occasionally, their computational expense is limited to specific instances, resulting in a minimal impact on short-term CPU usage. At lower frame rates, the reduced number of processed images naturally decreases LC queries, lowering the overall computational load.

Building on the ORB-SLAM3 base, **HFNet-SLAM** replaces classical feature extraction with learning-based methods (SuperPoint for local features and NetVLAD for global descriptors), resulting in higher resource utilization, particularly in memory usage. This increased demand stems from the reliance on deep network inference, which requires frequent CPU-GPU data transfers and additional database queries during LC. Even without LC, the overhead from learning-based feature extraction is evident, reflecting the computational cost of deep feature processing. Compared to ORB-SLAM3, the differences between LC configurations are more pronounced in terms of CPU utilization, likely due to the higher cost of global descriptor matching compared to the more efficient DBoW2-based approach.

**VINS-Fusion** utilizes an efficient frontend based on Shi-Tomasi corner detection with KLT tracking and, for loop detection, additionally extracts BRIEF descriptors for use with DBoW2. Loop correction is achieved via PGO. The computational results indicate that CPU usage strongly correlates with frame rate, increasing significantly at 30 FPS due to frequent local BA updates. Since BRIEF descriptors are only extracted when LC is enabled, LC introduces a slight additional CPU overhead, but its impact remains minimal. At lower frame rates, overall CPU usage drops significantly as fewer frames undergo local BA and LC queries. Despite the additional processing for LC, memory usage remains very low, highlighting the efficiency of PGO and KLT tracking, which avoid descriptor-based feature matching for pose estimation.

**OpenVINS** is extended with the LC module from VINS-Fusion and HLoc in a loosely coupled manner. While the additional extraction of BRIEF descriptors and PGO operations function similarly to VINS-Fusion, their impact on OpenVINS is significantly greater. CPU usage remains moderate and comparable for both VINS-Fusion and HLoc at 30 and 15 FPS but drops significantly at 5 FPS, indicating a strong correlation between computational load and frame rate, likely due to frequent feature extraction and LC candidate retrieval. When using HLoc as the LC module, CPU usage is slightly higher than with VINS-Fusion across all frame rates, likely due to the increased demand from deep network inference. This overhead arises from frequent CPU-GPU data transfers, particularly for global place recognition using NetVLAD and local feature matching with SuperPoint and SuperGlue, which contribute to the additional computational burden. In contrast, OpenVINS without LC exhibits significantly lower CPU demand across all frame rates, reflecting the efficiency of its frontend. This efficiency stems from the combination of Shi-Tomasi corner detection with KLT tracking, which enables lightweight feature processing, along with MSCKF-based state estimation that marginalizes past states to reduce computational load. Memory usage increases drastically when LC is enabled, as the LC thread independently retains additional data such as keyframes, descriptors, and pose graph information without efficiently pruning outdated states, since OpenVINS was not originally designed for LC integration. The impact is particularly pronounced with HLoc, which exhibits the highest memory consumption among all evaluated methods. This is likely due to the storage of high-dimensional descriptors from NetVLAD and SuperPoint, which are required for place recognition and geometric verification. Unlike VINS-Fusion's LC module, which relies on lightweight BRIEF descriptors, HLoc's learning-based approach significantly increases memory demand, with frame rate playing only a minor role in its overall footprint.

**Kimera** employs a Shi-Tomasi corner detector with KLT tracking in the frontend, while ORB features are extracted specifically for loop detection using DBoW2, and loop correction via PGO. Computational results show that CPU usage is highly dependent on frame rate, with a significant spike at 30 FPS when LC is enabled, reaching the highest utilization among all evaluated methods. This extreme CPU demand is likely due to frequent LC attempts and frequent PGO updates triggered by additional ORB extraction. At lower frame rates, CPU demand decreases, though LC still introduces noticeable overhead. Without LC, CPU usage is significantly lower, reflecting the reduced computational load when ORB extraction and DBoW2 searches are not required. Memory usage increases when LC is enabled but remains moderate compared to the CPU impact, suggesting that the primary bottleneck in Kimera is computational overhead rather than memory constraints.

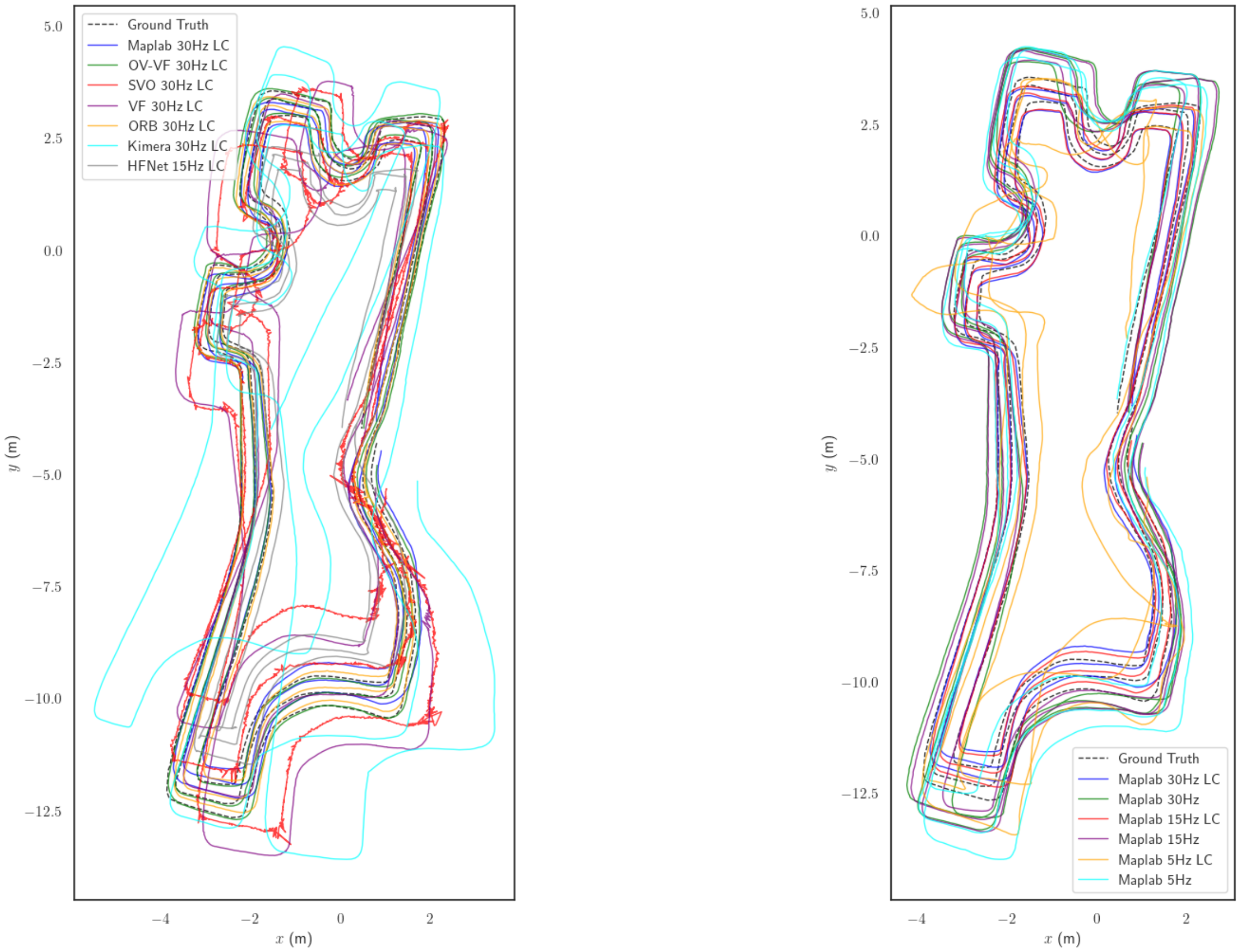

(a) SLAM Methods

(b) LC and FPS Configurations

**FIGURE 2** Trajectories for the *Garden Small Perimeter* scenario: (a) illustrates the comparison of the SLAM methods while (b) shows the differences in LC and FPS configurations of OpenVINS+Maplab.

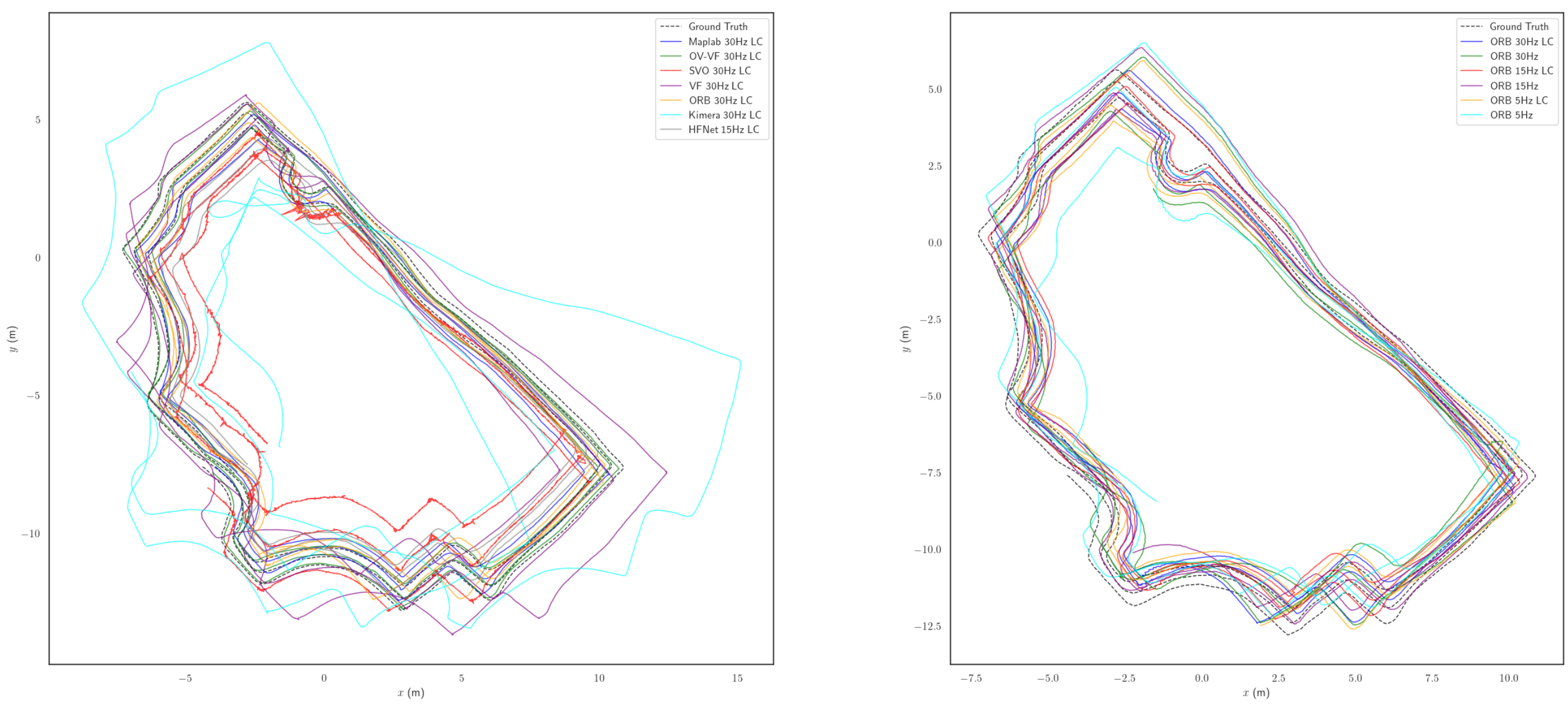

(a) SLAM Methods

(b) LC and FPS Configurations

**FIGURE 3** Trajectories for the *Garden Medium Perimeter* scenario: (a) illustrates the comparison of the SLAM methods while (b) shows the differences in LC and FPS configurations of ORB-SLAM3.

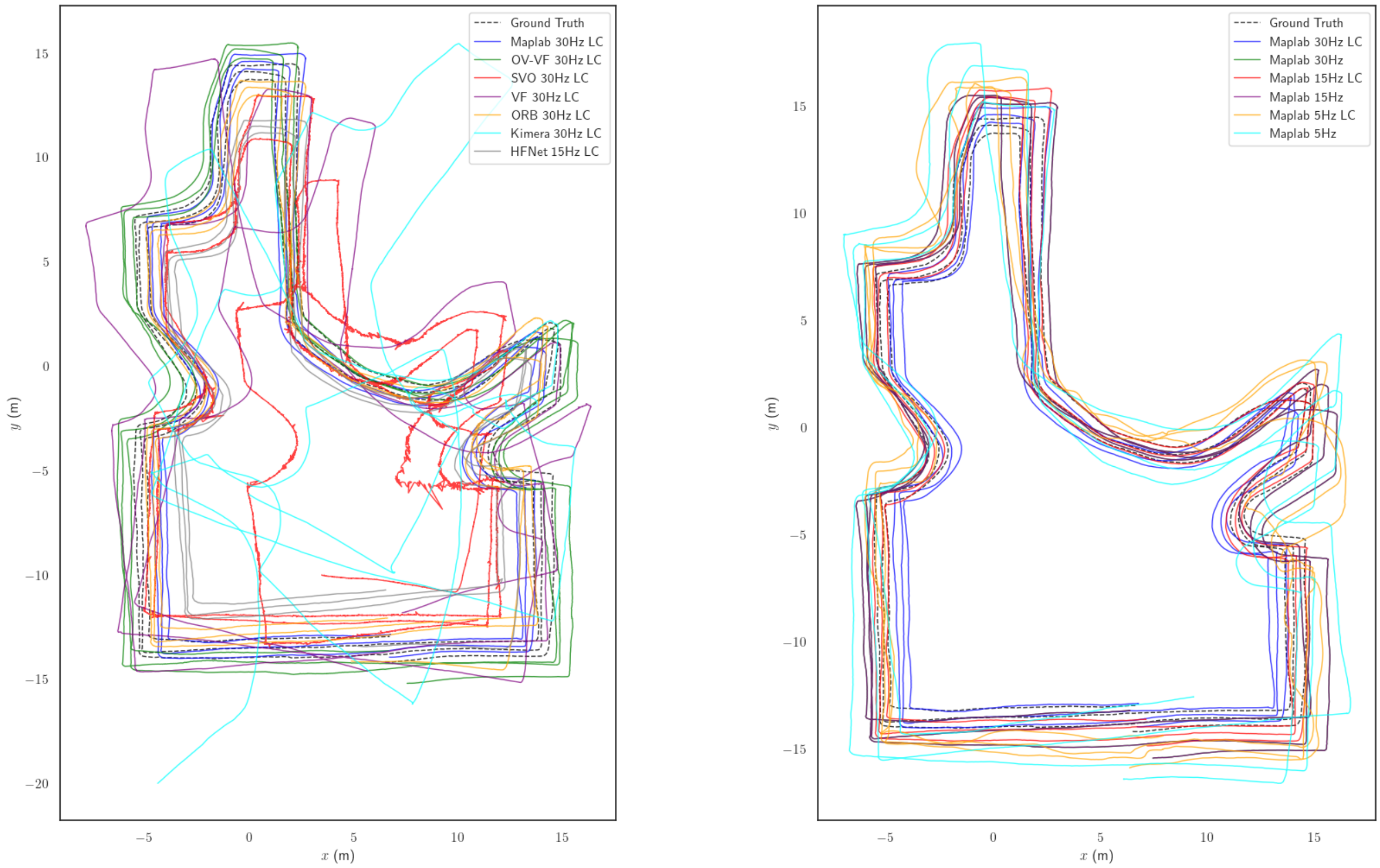

(a) SLAM Methods

(b) LC and FPS Configurations

**FIGURE 4** Trajectories for the *Garden Large Perimeter* scenario: (a) illustrates the comparison of the SLAM methods while (b) shows the differences in LC and FPS configurations of OpenVINS+Maplab.

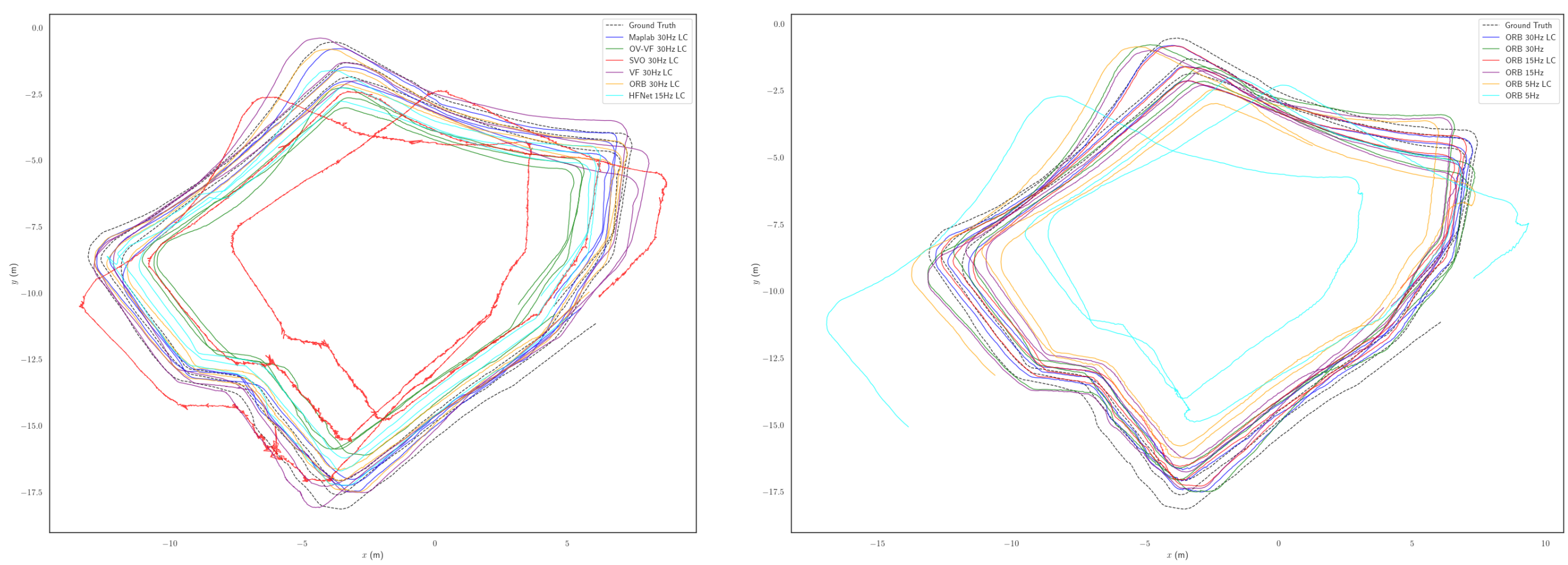

(a) SLAM Methods

(b) LC and FPS Configurations

**FIGURE 5** Trajectories for the *Park Perimeter* scenario: (a) illustrates the comparison of the SLAM methods while (b) shows the differences in LC and FPS configurations of ORB-SLAM3.

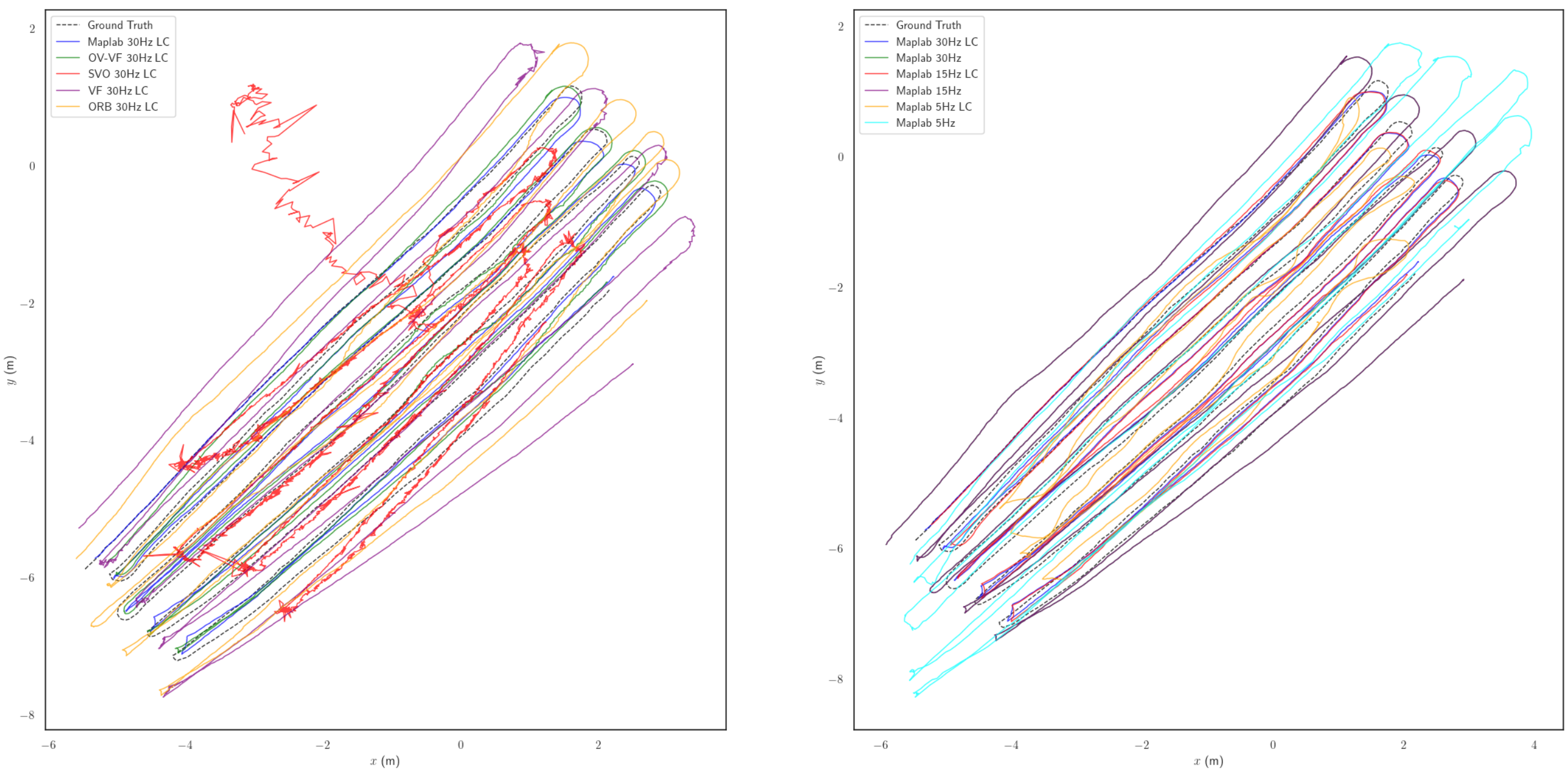

(a) SLAM Methods (b) LC and FPS Configurations

**FIGURE 6** Trajectories for the *Garden Medium Lane* scenario: (a) illustrates the comparison of the SLAM methods while (b) shows the differences in LC and FPS configurations of OpenVINS+Maplab.

**SVO Pro** employs a semi-direct approach, combining direct intensity-based methods with feature-based processing for pose estimation. LC is performed using DBoW2 with ORB feature extraction, while PGO is used for loop correction. Computational results show that CPU usage increases moderately when LC is enabled, primarily due to additional ORB extraction and DBoW2 queries. At 30 FPS, CPU utilization is significantly higher compared to lower frame rates, reflecting the increased computational demand from frequent keyframe processing and LC attempts. Without LC, CPU usage is significantly lower, indicating that LC contributes to a noticeable, though manageable, computational overhead. Memory usage remains low, with LC introducing only a slight increase. This suggests that SVO Pro efficiently manages its keyframes and map data, preventing excessive memory consumption. As the frame rate decreases, both CPU and memory usage drop significantly, reflecting the lower frequency of feature extraction, LC queries, and optimization updates.

Overall, the computational efficiency analysis highlights several key trends among the evaluated SLAM methods. Feature extraction strategies significantly influence resource consumption, with learning-based extractors such as SuperPoint and NetVLAD in HFNet-SLAM and OpenVINS+HLoc imposing higher computational costs due to GPU orchestration and frequent CPU-GPU data transfers. In contrast, classical extractors like ORB and BRIEF are computationally lighter. However, even among classical feature-based methods, ORB extraction and descriptor matching are generally more resource-intensive than Shi-Tomasi with KLT tracking, which rely on sparse optical flow for more efficient frame-to-frame tracking in real-time Visual-Inertial Odometry. The choice of backend optimization also affects resource usage, with methods employing local BA typically exhibiting higher CPU demand compared to filter-based approaches like OpenVINS, which leverages MSCKF. The marginalization strategy used in MSCKF helps maintain a lower computational and memory footprint. The manner in which LC is integrated into the system plays a crucial role in determining efficiency. Tightly coupled LC implementations, as seen in ORB-SLAM3 and VINS-Fusion, introduce minimal computational overhead, whereas loosely coupled configurations, such as OpenVINS with the VINS-Fusion LC module, significantly increase memory consumption due to less efficient keyframe and data management. Enabling LC necessitates additional feature extraction such as ORB or BRIEF, resulting in increased computational costs, especially at higher frame rates where LC operations occur more frequently. The cost of graph optimization also varies, with global BA being significantly more resource-intensive since it jointly optimizes a large set of keyframes and landmarks. In contrast, PGO only refines keyframe poses based on their relative transformations, making it computationally more efficient as it avoids direct optimization of 3D landmark positions. Finally, frame rate strongly influences computational load, as higher FPS increases the number of processed images per second, amplifying both LC feature extraction and pose graph updates, leading to a proportional rise in computational demand.

TABLE 7 CPU and Memory Usage in *Garden Medium Perimeter* scenario with and without LC at 30 FPS, 15 FPS, and 5 FPS. All reported results involve the publicly available open-source implementations. An 'x' indicates that the method failed during execution, and no values are reported, as they would not be comparable to methods that successfully completed the sequence.

| Method | FPS | LC | CPU Usage [%] Min | CPU Usage [%] Mean | CPU Usage [%] Median | CPU Usage [%] Max | Memory Usage [GB] Min | Memory Usage [GB] Mean | Memory Usage [GB] Median | Memory Usage [GB] Max |
|---|---|---|---|---|---|---|---|---|---|---|
| ORB-SLAM3 (M-I) | 30 | On | 303.80 | 419.50 | 428.40 | 521.20 | **0.59** | 1.57 | 1.64 | 2.18 |
| | | Off | 301.90 | 417.85 | 425.80 | 462.30 | 0.60 | 1.52 | 1.60 | 2.00 |
| | 15 | On | 232.10 | 325.35 | 332.90 | 366.90 | **0.59** | 1.48 | 1.54 | 2.11 |
| | | Off | **229.30** | **325.06** | **331.10** | **361.10** | 0.60 | **1.42** | **1.48** | **1.95** |
| | 5 | On | x | x | x | x | x | x | x | x |
| | | Off | x | x | x | x | x | x | x | x |
| HFNet-SLAM | 30 | On | 91.60 | 483.81 | 484.00 | 998.00 | 2.28 | 4.62 | 4.75 | 6.39 |
| | | Off | 92.70 | 395.24 | 398.40 | 881.90 | 2.28 | 4.42 | 4.46 | 6.13 |
| | 15 | On | 47.90 | 405.09 | 404.70 | 706.10 | **2.27** | 4.34 | 4.38 | 6.23 |
| | | Off | 46.80 | 358.38 | 354.50 | 653.40 | 2.28 | 4.16 | 4.36 | 5.51 |
| | 5 | On | **15.90** | 334.78 | 354.70 | 670.20 | **2.27** | 3.86 | 3.85 | 5.52 |
| | | Off | **15.90** | **308.68** | **304.80** | **503.20** | **2.27** | **3.70** | **3.81** | **4.88** |
| ORB-SLAM3 (S-I) | 30 | On | 371.10 | 403.02 | 401.20 | 475.50 | 0.52 | 0.96 | 0.96 | 1.39 |
| | | Off | 369.30 | 400.73 | 399.40 | 451.30 | 0.52 | 0.93 | 0.93 | 1.34 |
| | 15 | On | 281.50 | 315.23 | 315.40 | 350.90 | 0.52 | 0.97 | 0.98 | 1.36 |
| | | Off | 282.00 | 309.97 | 309.00 | 358.90 | 0.51 | 0.94 | 0.94 | 1.32 |
| | 5 | On | **227.10** | **246.91** | 247.80 | **268.70** | 0.52 | **0.79** | **0.79** | **1.06** |
| | | Off | 227.30 | 247.47 | **247.10** | 273.10 | **0.51** | 0.82 | 0.82 | 1.13 |
| VINS-Fusion | 30 | On | **32.80** | 266.84 | 270.90 | 321.40 | **0.09** | 0.11 | 0.12 | 0.12 |
| | | Off | 58.60 | 243.23 | 236.90 | 365.80 | **0.09** | 0.11 | **0.10** | 0.14 |
| | 15 | On | 109.40 | 138.73 | 139.40 | 173.10 | **0.09** | **0.10** | **0.10** | 0.11 |
| | | Off | 108.40 | 129.86 | 129.00 | 161.00 | **0.09** | **0.10** | **0.10** | 0.11 |
| | 5 | On | 37.80 | 48.97 | 48.80 | 59.70 | **0.09** | **0.10** | **0.10** | 0.11 |
| | | Off | 35.90 | **46.72** | **46.80** | **56.80** | **0.09** | **0.10** | **0.10** | **0.10** |
| OpenVINS + VINS-Fusion + HLoc | 30 | VF | 7.90 | 139.51 | 135.30 | 235.20 | 0.50 | 3.21 | 3.23 | 5.89 |
| | | HLoc | 12.90 | 169.67 | 175.10 | 207.90 | 4.52 | 5.91 | 5.93 | 7.21 |
| | | Off | **2.00** | 65.00 | 65.80 | 95.70 | **0.08** | **0.29** | **0.29** | **0.29** |
| | 15 | VF | 23.70 | 141.14 | 137.40 | 242.30 | 0.61 | 3.21 | 3.22 | 5.89 |
| | | HLoc | 12.90 | 168.80 | 173.30 | 204.90 | 4.47 | 5.84 | 5.86 | 7.13 |
| | | Off | 10.00 | 55.52 | 55.80 | 87.70 | 0.22 | **0.29** | **0.29** | 0.30 |
| | 5 | VF | 13.80 | 74.94 | 74.75 | 107.90 | 0.58 | 1.57 | 1.59 | 2.51 |
| | | HLoc | 6.00 | 79.37 | 81.60 | 142.40 | 4.43 | 5.15 | 5.15 | 5.73 |
| | | Off | 5.90 | **41.75** | **43.70** | **48.70** | 0.19 | 0.32 | 0.33 | 0.34 |
| Kimera | 30 | On | 66.70 | 638.89 | 625.65 | 1490.00 | 0.56 | 1.37 | 1.38 | 2.07 |
| | | Off | 69.30 | 163.16 | 167.20 | 190.20 | 0.29 | 0.56 | 0.57 | 0.62 |
| | 15 | On | 37.90 | 443.41 | 354.30 | 1111.30 | 0.68 | 1.28 | 1.29 | 1.82 |
| | | Off | 39.70 | 125.49 | 128.70 | 173.80 | 0.21 | 0.35 | 0.35 | 0.39 |
| | 5 | On | 15.00 | 356.38 | 122.00 | 1410.30 | 0.56 | 1.15 | 1.15 | 1.59 |
| | | Off | **14.00** | **71.89** | **70.90** | **122.50** | **0.17** | **0.30** | **0.30** | **0.34** |
| SVO Pro | 30 | On | 60.70 | 289.07 | 295.90 | 351.50 | **0.10** | 0.72 | 0.75 | 1.29 |
| | | Off | **6.00** | 163.11 | 166.40 | 203.10 | **0.10** | 0.18 | 0.18 | 0.18 |
| | 15 | On | 23.90 | 162.30 | 162.50 | 226.20 | **0.10** | 0.51 | 0.50 | 0.85 |
| | | Off | 28.90 | 98.51 | 100.80 | 125.60 | **0.10** | 0.18 | 0.18 | 0.18 |
| | 5 | On | 10.00 | 52.44 | 52.80 | 71.70 | **0.10** | 0.29 | 0.29 | 0.39 |
| | | Off | 11.00 | **26.13** | **26.90** | **31.90** | **0.10** | **0.17** | **0.17** | **0.17** |

## 5 Discussion

On the impact of LC on SLAM algorithms, our study reveals that, while LC broadly enhances localization accuracy, its influence on computational demands diverges across different systems (see Figure 7). This divergence highlights an essential strategic balance that must be struck between accuracy enhancement and the management of computational resources. Further, the distinction in performance between driving scenarios, i.e., *Perimeter* and *Lane*, refines our understanding of LC's effects, underlining the importance of context in evaluating SLAM system performance.

**ORB-SLAM3** with LC consistently improved localization accuracy across different scenarios, maintaining low ATE values for both *Perimeter* and *Lane* scenarios. Higher frame rates (30 FPS and 15 FPS) significantly enhanced performance, particularly with LC, while at 5 FPS, the *Lane* scenario performance declined notably. The stereo-inertial setup generally provided better accuracy across environments, whereas the mono-inertial setup struggled in larger environments like *Garden Medium* and *Garden Large*, though LC improved results, especially in the *Perimeter* scenario. Computationally, ORB-SLAM3 remained resource-intensive, with the mono-inertial setup consuming slightly more CPU and about 50% more memory than the stereo-inertial version. Memory usage was otherwise stable across frame rates and LC settings. Overall, ORB-SLAM3 benefits from higher frame rates and LC but at the cost of increased resource consumption, with the stereo-inertial setup offering more reliable performance.

**HFNet-SLAM** performed best at lower frame rates, particularly in the *Perimeter* scenario, where LC improved accuracy, while *Lane* remained stable with minimal LC impact. Computationally, it had the highest resource demands, exceeding ORB-SLAM3 in CPU and memory usage. LC increased CPU load, while lower frame rates reduced it. Memory consumption was about three times higher than ORB-SLAM3 mono-inertial, with GPU VRAM usage remaining constant at 502 MB. The trade-off between localization accuracy and computational expense is not well balanced, but HFNet-SLAM is more stable compared to ORB-SLAM3 mono-inertial and could be a viable solution where high computational demand is manageable and no stereo image is provided.

**VINS-Fusion** demonstrated consistent improvements in localization accuracy with LC, particularly at 30 FPS and 15 FPS. However, the overall localization accuracy was at a medium to low level, struggling especially in the *Lane* scenario and low frame rates. VINS-Fusion exhibited moderate CPU usage, which declined at lower frame rates, and very low memory demands, making it efficient for environments with limited resources. Despite its resource efficiency, VINS-Fusion struggles at lower frame rates and in complex scenarios, suggesting that while it is a resource-efficient option its accuracy may not be adequate for dynamic or highly complex environments.

**OpenVINS+VINS-Fusion** and **OpenVINS+HLoc** both enhance OpenVINS with LC but exhibit distinct trade-offs in accuracy and computational efficiency. OpenVINS+VINS-Fusion maintained stable performance with LC, particularly in *Perimeter* at higher frame rates, though it struggled in *Lane*. OpenVINS+HLoc, while less accurate in *Perimeter*, was more robust in *Lane*, providing stable results across frame rates where OpenVINS+VINS-Fusion and OpenVINS without LC mostly failed. Computationally, both methods had moderate CPU usage at 30 and 15 FPS, dropping significantly at 5 FPS. OpenVINS+HLoc required slightly more CPU due to deep learning-based feature extraction and frequent CPU-GPU data transfers. However, memory consumption was drastically higher with LC, particularly for OpenVINS+HLoc, making it the most memory-intensive SLAM method. The use of LC led to a substantial increase in computational demand, with CPU usage nearly doubling and memory usage increasing by an order of magnitude compared to running OpenVINS without LC. This trade-off makes LC best suited for environments with ample computational resources, as its increased demand may be impractical for resource-constrained applications. Among the LC-enabled configurations, OpenVINS+VINS-Fusion offers the best balance between accuracy and efficiency, while OpenVINS without LC remains a viable option for low-computation scenarios, delivering good results with minimal resource usage.

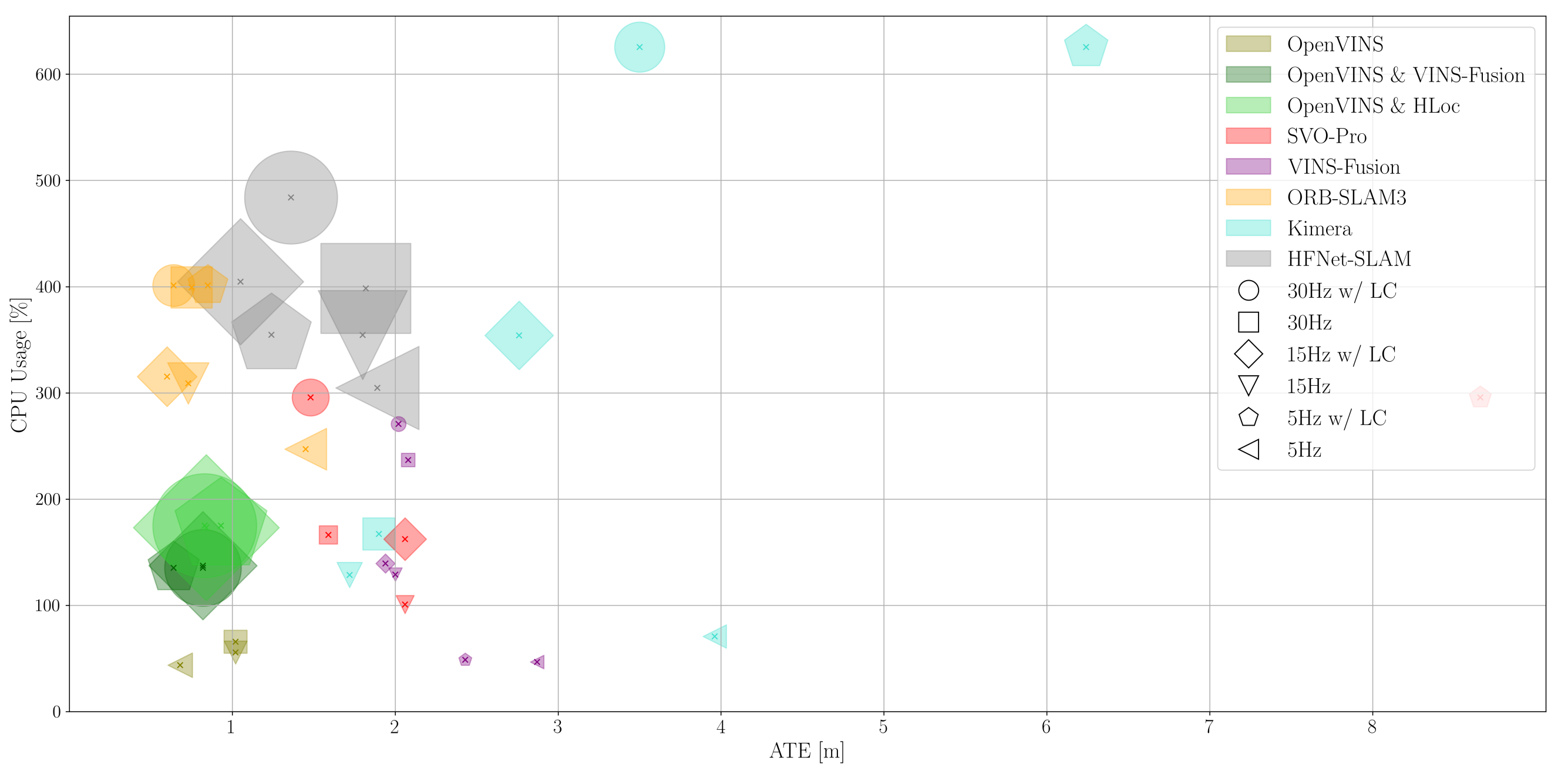

**FIGURE 7** Comparison of RMSE ATE versus CPU utilization for the SLAM methods and their configurations, considering LC and FPS, in the *Garden Medium Perimeter* scenario. Symbol size corresponds proportionally to memory consumption.

**OpenVINS+Maplab** consistently performed well with LC, especially at higher frame rates, maintaining low ATE values in the *Perimeter* scenario and showing improved accuracy overall. Lower frame rates increased ATE, particularly in complex scenarios, emphasizing the importance of high frame rates for accuracy. The *Lane* scenario showed improvement with LC but experienced greater difficulties in large environments and at lower frame rates. Operating offline, Maplab avoids real-time computational constraints, ideal for applications with post-processing capabilities. This method is suitable for environments where high computational demand can be managed in non-real-time settings.

**Kimera** exhibits high variance in localization accuracy, demonstrated by elevated ATE values across various environments and particularly in the Lane scenario at low frame rates. CPU usage was notably high, especially with LC enabled at 30 FPS, while memory usage ranged from moderate to high. This suggests that Kimera's high resource utilization does not translate into equivalent accuracy gains, making it less suitable for applications with constrained computational resources.

**SVO Pro** showed improved performance with LC, particularly at higher frame rates, demonstrating reasonable localization accuracy in the *Perimeter* scenario but struggling more with the *Lane* scenario. CPU usage was at a medium level and significantly increased with LC, while memory usage remained low. SVO Pro benefits from higher frame rates and LC for optimal accuracy, making it suitable for scenarios where medium computational loads are acceptable and higher frame rates can be maintained. However, due to its moderate accuracy and increased computational demands with LC, it may not be ideal for resource-constrained environments.

**AirSLAM** showed poor localization accuracy, with slight improvements from LC. ATE increased with environment size, while RPE remained moderate to poor. Lower frame rates improved ATE by reducing drift accumulation and redundant keyframes, enhancing global trajectory consistency, but worsened RPE due to larger gaps between frames affecting short-term pose stability. Conversely, higher frame rates improved RPE through smoother motion tracking but increased ATE due to more frequent drift accumulation. As AirSLAM applies LC post-processing, no computational analysis was conducted.

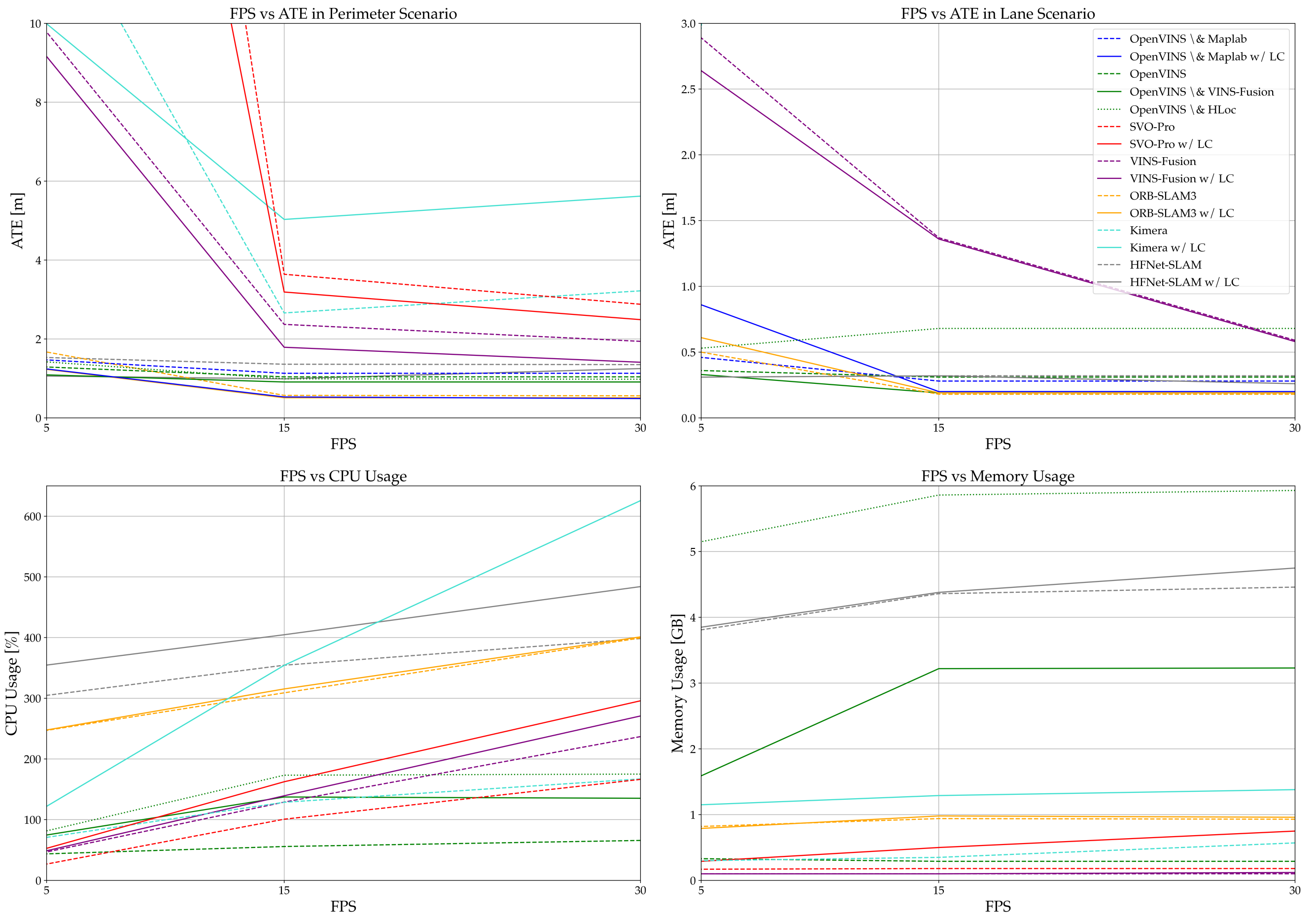

**FIGURE 8** Comparison of SLAM methods regarding ATE in Perimeter and Lane driving scenarios, as well as CPU and memory usage under different FPS, with and without LC.

Figure 8 visualizes the main findings of our study regarding the effects of LC and varying FPS on the ATE, CPU, and memory usage. ORB-SLAM3 in stereo-inertial setup with LC stands out as the best option. It provides superior precision at higher frame rates, making it ideal for environments that can support the necessary computational demands. OpenVINS+VINS-Fusion is an effective choice for unstructured outdoor settings with ample computational capacity. With LC, it maintains low ATE and demonstrates stable performance, making it suitable for detailed field analysis and large-scale outdoor data collection where high memory usage can be accommodated. Additionally, without LC, OpenVINS+VINS-Fusion achieves good accuracy results with a very low computational footprint, making it a suitable choice for embedded solutions with limited computational resources. OpenVINS+Maplab is particularly suited for applications that can handle post-processing. Its ability to offer superior localization accuracy without real-time computational constraints makes it perfect for scenarios where a detailed offline analysis of unstructured outdoor data is needed.

The comparison between the *Perimeter* and *Lane* scenarios reveals distinct differences in performance across various SLAM algorithms. Notably, the Lane scenario exhibits significant variability, displaying high accuracy and superior localization precision in certain environments and methods, such as ORB-SLAM3, OpenVINS+VINS-Fusion, and OpenVINS+Maplab. This suggests a potential for exceptional performance under favorable conditions. However, the same methods also encounter severe challenges in specific environments such as the *Garden Small*, *Medium*, and *Large*, where localization occasionally fails completely, indicating the methods' inability to handle the scenario reliably and lack of robustness. Overall, while the *Lane* scenario shows potential for high precision, its performance is notably inconsistent, making it less robust compared to the *Perimeter* scenario. This inconsistency can be attributed to the demanding nature of the *Lane* scenario, which includes multiple 180° turns. These maneuvers are likely problematic for localization, often leading to drift and contributing to the varied results observed across different SLAM implementations.

## 6 Conclusion

In our investigation of the impact of LC on Visual-Inertial SLAM systems in natural unstructured environments, we thoroughly evaluated traditional as well as learning-based methods in terms of localization accuracy and resource usage.

ORB-SLAM3 maintains high localization accuracy, with LC slightly improving it without introducing substantial computational overhead. However, the method requires considerable computational resources at all frame rates, making it suitable for high-precision tasks in resource-rich environments. HFNet-SLAM achieves stable localization accuracy, particularly at lower frame rates, but exhibits high computational demands, making it suitable only for scenarios where higher resource consumption is acceptable and no stereo images are available. VINS-Fusion offers moderate localization improvements with LC and struggles at lower frame rates, but its efficient CPU and memory usage makes it viable for resource-constrained applications where high localization accuracy is not required. OpenVINS+VINS-Fusion performs well, with LC providing a slight accuracy improvement at the cost of significant computational overhead. OpenVINS+HLoc achieves slightly lower localization accuracy while demanding even more computational resources, making it less efficient than OpenVINS+VINS-Fusion. Meanwhile, OpenVINS without LC remains a viable choice for low-computation scenarios, offering good results with minimal resource usage. OpenVINS+Maplab excels in post-processing scenarios, providing high localization accuracy without real-time computational constraints, ideal for detailed unstructured outdoor data analysis and long-term monitoring tasks. Kimera faces challenges with localization accuracy and high CPU usage with LC, making it less suitable for resource-constrained applications. SVO Pro benefits from LC and higher frame rates, maintaining reasonable localization accuracy but struggling with increased computational demands at lower frame rates. AirSLAM demonstrates poor localization accuracy, particularly in larger environments, with only slight improvements from LC, making it impractical for the presented scenarios.

## ACKNOWLEDGMENTS

We thank Joshua Uhl for his support during the data capturing and implementation phase and Frank Holzmüller and Manuel Kaiser for their initial support of the project. We also acknowledge Sabrina Kaniewski for her valuable assistance with reviewing and editing.

## ORCID

*Fabian Schmidt*
*Constantin Blessing*
*Markus Enzweiler*
*Abhinav Valada*